\documentclass{article}

\usepackage{amsmath,amsfonts}
\usepackage{iclr2027_conference,times}
\usepackage{hyperref}
\usepackage{multirow}
\usepackage{booktabs}
\usepackage{graphicx}
\usepackage{tabularx}
\usepackage{wrapfig}
\usepackage{caption}

\title{\centering Memory as the Backbone: \\A Dual-Store Associative Memory Architecture for Visual Recognition\vspace{1em}}
\iclrfinalcopy 

\author{%
\makebox[\dimexpr\textwidth-2\tabcolsep\relax][c]{%
\begin{tabular}{@{}c@{}}
{\normalsize\bfseries
Ruizhi Wang\textsuperscript{1}\thanks{Corresponding author: ruizhi.wang@tuwien.ac.at}
\quad
Xiangtao Wang\textsuperscript{1}
\quad
Bayar Menzat\textsuperscript{1}
}
\\[0.6em]
{\normalsize\bfseries
Jianfeng Wang\textsuperscript{2}
\quad
Thomas Lukasiewicz\textsuperscript{1,2}
}
\\[1em]
{\normalfont
\textsuperscript{1}Faculty of Informatics,
Vienna University of Technology, Austria}
\\[0.2em]
{\normalfont
\textsuperscript{2}Department of Computer Science,
University of Oxford, United Kingdom}
\end{tabular}%
}%
}

\begin{document}
\maketitle
\pagestyle{plain}
\thispagestyle{plain}

\begin{abstract} 
We investigate whether explicit memory can serve as the organizing principle of a general-purpose visual backbone. Inspired by the interaction between working and long-term memory in human cognition, we introduce \textbf{DuSAM} (\textbf{Du}al-\textbf{S}tore \textbf{A}ssociative \textbf{M}emory), a hierarchical memory-native visual backbone coupling image-local working memory (WM) with a fixed-capacity persistent semantic memory (PM). WM organizes local visual evidence into compact, spatially anchored states that are updated and transported across stages, then read back to refine dense features. PM consolidates semantic evidence from training images and annotated regions into prototypes, whose retrieved content enriches the same working states. With these memory operations in place of self-attention or state-space blocks, DuSAM achieves competitive performance relative to representative, similarly sized backbones on ImageNet-1K classification, ADE20K semantic segmentation, and COCO object detection and instance segmentation. Ablations show contributions from both stores, with PM particularly benefiting class groups with fewer annotated training pixels. At $2048\times2048$, the complete DuSAM-based UPerNet segmentor achieves $2.36\times$ the throughput of the evaluated DeiT-S/16 counterpart while using $59.5\%$ less GPU memory, and $1.24\times$ the throughput of the linear-complexity VMamba-T. Memory analyses reveal spatially selective WM access and semantic associations between retrieved prototypes and training regions. Class-targeted removal and restoration further demonstrate that category-specific experience can be replenished through state updates without changing network parameters. These results establish explicit memory as a viable organizing principle for visual backbones and provide a foundation for jointly designing contextual computation and addressable semantic memory.
\end{abstract}

\section{Introduction}
\label{sec:introduction}
Visual backbones have evolved around different ways of exchanging information. Convolutional networks build context through local, weight-shared operations \citep{DBLP:conf/cvpr/HeZRS16}. Vision Transformers organize interactions through token-based self-attention \citep{DBLP:conf/iclr/DosovitskiyB0WZ21}, while hierarchical variants extend these interactions to multi-scale representations and dense prediction \citep{DBLP:conf/iccv/LiuL00W0LG21,DBLP:conf/iccv/WangX0FSLL0021}. More recently, visual state-space models use selective scans and state recurrence to propagate information \citep{DBLP:conf/icml/ZhuL0W0W24,DBLP:conf/nips/LiuTZYX0YJ024}. In this work, we ask: \emph{can a general-purpose visual backbone be built around explicit memory, using memory writes and reads to organize the exchange of visual information?}

Explicit memory organizes information as addressable states that can be stored, updated, and retrieved. Existing approaches have used memory to extend language-model context \citep{DBLP:conf/iclr/WuRHS22}, supply cross-image reference features for contrastive learning \citep{DBLP:conf/iccv/AlonsoSFMM21}, and preserve evidence over long video sequences \citep{DBLP:conf/eccv/ChengS22}. For a backbone built around memory, both access and state organization matter. Associative retrieval provides content-based access to stored representations, with modern Hopfield formulations making this operation differentiable \citep{DBLP:conf/iclr/RamsauerSLSWGHA21}. Such a retrieval rule alone, however, does not determine what to retain or how these states should support a spatial hierarchy.

Visual recognition depends on both the spatial context of a particular scene and semantic regularities shared across scenes. The former must remain grounded in the current image and evolve with its features; the latter should remain available beyond a single input. We draw inspiration from working-memory accounts in human cognition, where a limited-capacity workspace integrates current information with content retrieved from long-term memory \citep{baddeley2000episodic}. This suggests a design principle: maintain current-image context and accumulated semantic experience in separate stores, while bringing them together during visual processing.

Following this principle, we introduce \textbf{DuSAM} (\textbf{Du}al-\textbf{S}tore \textbf{A}ssociative \textbf{M}emory), a brain-inspired hierarchical visual backbone. Image-local working memory (WM) gathers locally extracted features into compact, spatially anchored states that are transported across stages and updated with new evidence. Persistent semantic memory (PM) consolidates evidence from training images or annotated regions into a fixed-capacity prototype bank shared across the deeper stages. At these stages, image tokens retrieve prototypes through content similarity and predicted semantic addresses. The retrieved content enriches the working states, which are then read back to update dense features. WM is rebuilt for each image, whereas PM persists across training samples and is fixed at inference.

This organization has two useful consequences. Image locations exchange context through compact working states and access a bounded prototype bank, so memory interactions scale linearly with token count for fixed widths and capacities. In addition, part of the semantic experience used in prediction remains accessible as explicit content. Retrieved prototypes can be associated with training-region exemplars, and selected entries can be replenished without weight updates.

We evaluate DuSAM on ImageNet-1K classification, ADE20K semantic segmentation, and COCO object detection and instance segmentation. It achieves 79.1\% top-1 accuracy, $43.3$ single-scale and $44.5$ multi-scale mIoU, and $42.6$ box AP and $38.9$ mask AP, respectively, demonstrating competitive performance relative to representative backbones of similar size. At $2048\times2048$, the complete DuSAM-based UPerNet segmentor achieves $2.36\times$ the throughput of the evaluated DeiT-S/16 counterpart while using $59.5\%$ less GPU memory, and $1.24\times$ that of the linear-complexity VMamba-T.

Ablations show that both stores contribute to dense prediction, with PM particularly benefiting class groups with fewer annotated training pixels. Working-state analyses reveal spatially differentiated reads, while prototype visualizations expose the visual associations of retrieved semantic content. Category-specific experience can also be replenished without re-optimizing the network: writing five annotated training regions restores up to $93\%$ of the IoU lost after target-class memory removal. 
Together, these findings demonstrate the viability of explicit memory as an organizing principle for visual backbones. They provide a foundation for studying visual architectures that jointly organize contextual computation and semantic experience through addressable states.

\section{Related Work}
\label{sec:related_work}

\textbf{Visual backbone architectures.}
Visual backbone research explores new architectures for representation learning. Convolutional backbones improve feature extraction through residual learning, channel recalibration, and modernized blocks \citep{DBLP:conf/cvpr/HeZRS16,DBLP:conf/cvpr/HuSS18,DBLP:conf/cvpr/0003MWFDX22}. Attention-based approaches introduce content-dependent token interactions \citep{DBLP:conf/iclr/DosovitskiyB0WZ21,DBLP:conf/icml/TouvronCDMSJ21}, while pyramids and shifted windows support multi-scale representations \citep{DBLP:conf/iccv/LiuL00W0LG21,DBLP:conf/iccv/WangX0FSLL0021}. MLP and pooling-based alternatives learn competitive representations without attention \citep{DBLP:conf/nips/TolstikhinHKBZU21,DBLP:conf/cvpr/YuLZSZWFY22}, while state-space, recurrent, and hybrid models propagate context through state updates~\citep{DBLP:conf/icml/ZhuL0W0W24, DBLP:conf/nips/LiuTZYX0YJ024, DBLP:conf/iclr/AlkinBPHB25, DBLP:conf/nips/ShiDX24, DBLP:conf/cvpr/HatamizadehK25}. Recent work refines state fusion~\citep{DBLP:conf/iclr/Xiao0ZM025}, questions the need for sequence mixers~\citep{DBLP:conf/cvpr/0001W25}, or builds on heat conduction~\citep{DBLP:conf/cvpr/WangLT0WY25}.
Despite these computational differences, cross-image semantic knowledge remains primarily in learned parameters. DuSAM represents part of this knowledge as explicit memory content, supporting prototype-level interpretation of the semantic experience used in recognition.

\textbf{Associative and neural memory.}
Neural memory can separate stored information from the parameters that process it. Differentiable read--write systems retain and manipulate intermediate information \citep{DBLP:journals/nature/GravesWRHDGCGRA16}, while modern Hopfield formulations link content-addressed retrieval to softmax attention \citep{DBLP:conf/iclr/RamsauerSLSWGHA21}. External stores extend sequence models' access to past representations, while multi-store designs manage retention and forgetting \citep{DBLP:conf/iclr/WuRHS22,DBLP:conf/icml/ParkB24}. AiT~\citep{DBLP:conf/cvpr/SunOW0K25} adds associative memory to Transformer blocks and V-HMN~\citep{wang2026vision} uses local/global prototype retrieval, neither demonstrating transfer to detection or segmentation; LookupViT~\citep{DBLP:conf/eccv/KonerJJTP24} exchanges information between dense and compressed tokens within a Transformer. These studies leave open how transient image context and accumulated semantic experience should cooperate across feature scales, which DuSAM addresses through associative writes, cross-stage state transport, and persistent semantic retrieval.

\textbf{Memory in visual perception.}
Memory supplies visual evidence beyond a single input. In semantic segmentation, class-wise banks provide cross-image references for contrastive learning \citep{DBLP:conf/iccv/AlonsoSFMM21}, while MCIBI~\citep{DBLP:conf/iccv/JinGY0WWS21} uses predicted class probabilities to retrieve dataset-level context and enhance features produced by an existing segmentation backbone. For video, short and long-term stores preserve evidence across frames \citep{DBLP:conf/eccv/ChengS22}; object-level reading and consolidation of past activations improve the use of temporal context \citep{DBLP:conf/cvpr/ChengOPLS24,DBLP:conf/icml/BalazevicSP0KH24}. These task-specific uses augment features supplied by an existing backbone. DuSAM instead builds the feature hierarchy around memory computation, yielding a general-purpose backbone for classification, detection, and segmentation.

\section{Methodology}
\label{sec:Methodology}

\begin{figure}[t]
    \centering
    \includegraphics[width=\linewidth]{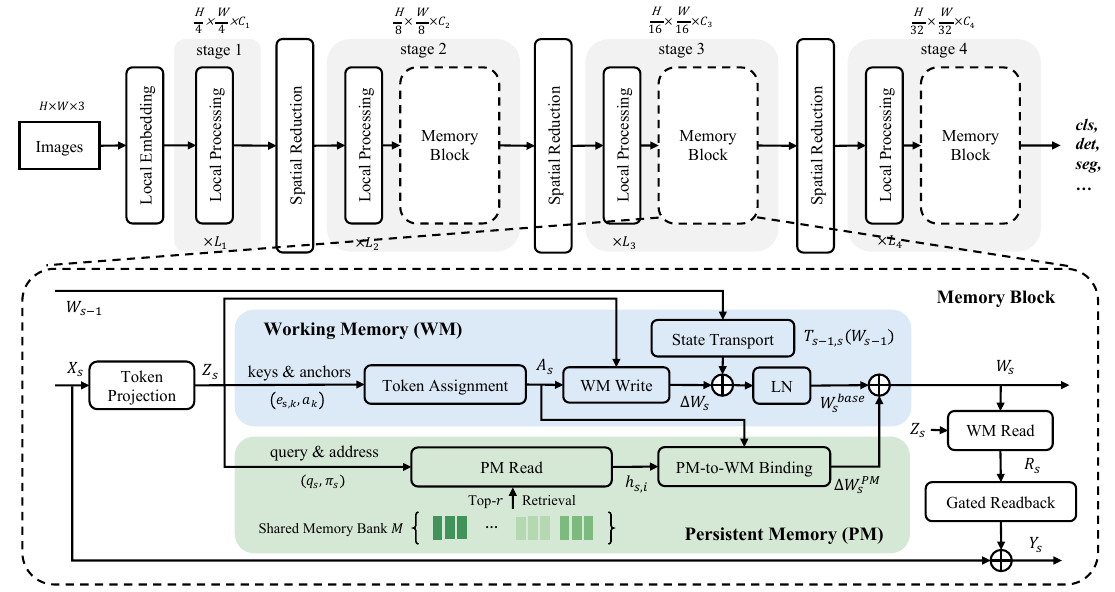}
    \caption{Overview of DuSAM. 
    \emph{Upper:} the four-stage backbone. \emph{Lower:} a Memory Block. WM Write and State Transport construct the workspace, PM-to-WM Binding integrates retrieved semantics, and WM Read with Gated Readback updates dense features.}
    \label{fig:architecture}
\end{figure}

DuSAM organizes visual computation through two complementary stores: Working Memory (WM) gathers context within the current image, while Persistent Memory (PM) retains semantic experience across training images. As shown in Figure~\ref{fig:architecture}, local embedding and successive spatial reductions form a four-stage hierarchy. Stage~1 uses local processing alone; Stages~2--4 additionally contain one Memory Block, with Stages~3--4 accessing a shared PM. See Appendix~\ref{app:implementation} for details.

\subsection{Image-Local Working Memory}
\label{sec:wm}
\textbf{Spatially anchored state construction.}
WM enables distant image locations to exchange context through $K$ working states. Fixed anchors $a_k$ at multi-resolution grid centers provide spatial references, while learned assignment keys $e_{s,k}$ guide content collection. A pointwise projection $E_s$ maps the local features $X_s\in\mathbb{R}^{N_s\times C_s}$ to token representations $Z_s=E_s(X_s)$, where $N_s=H_sW_s$. For a token at position $p_i$, its affinity with state $k$ combines content similarity and spatial proximity:
\begin{equation}
 u_{s,ik}
 =
 \frac{\langle\widehat z_{s,i},\widehat e_{s,k}\rangle}{\tau_w}
 -
 \lambda_{\mathrm{sp}}\lVert p_i-a_k\rVert_2^2,
 \label{eq:wm-affinity}
\end{equation}
where $\widehat{\cdot}$ denotes $\ell_2$ normalization, and $\tau_w$ and $\lambda_{\mathrm{sp}}$ control content temperature and spatial weighting, respectively. To distribute token contributions more evenly across states, we normalize these affinities with a fixed number of Sinkhorn iterations~\citep{DBLP:conf/nips/Cuturi13}, obtaining an approximately balanced assignment $A_s\in\mathbb{R}^{N_s\times K}$. We then form working states by weighted token aggregation:
\begin{equation}
 \bar A_{s,ik}
 =
 \frac{A_{s,ik}}{\max(\sum_j A_{s,jk},\epsilon_w)},
 \qquad
 \Delta w_{s,k}
 =
 \operatorname{LN}_0\!\left(
 \sum_i\bar A_{s,ik}z_{s,i}
 \right),
 \label{eq:wm-write}
\end{equation}
where $\operatorname{LN}_0$ denotes channel-wise layer normalization.

\textbf{Cross-stage update and dense readback.}
To preserve context across stages, we project previous states to the current width with a learned linear map $T_{s-1,s}$ and add new evidence:
\begin{equation}
 W_s^{\mathrm{base}}
 =
 \operatorname{LN}\!\left(
 T_{s-1,s}(W_{s-1})+\Delta W_s
 \right),
 \qquad s\in\{3,4\}.
 \label{eq:wm-transport}
\end{equation}
Stage~2 initializes $W_2=\Delta W_2$ without transport or PM. At deeper stages, PM enriches the base states into $W_s$, as described below. Each token then retrieves context by content similarity and incorporates it through a gated residual:
\begin{equation}
 B_s=\operatorname{softmax}_{k}\!\left(
 \widehat Z_s\widehat W_s^{\mathsf T}/\tau_w
 \right),
 \quad
 R_s=B_sW_s,
 \quad
 Y_s=X_s+\sigma(g_s)\odot O_s(R_s).
 \label{eq:wm-readback}
\end{equation}
$O_s$ is an output projection and $\sigma(g_s)$ a learned channel gate. The write weights $A_s$ determine what each state collects, whereas the read weights $B_s$ determine what each token retrieves. The block outputs both refined features $Y_s$ and updated working states $W_s$.

For fixed $K$, matching and aggregation through working states cost $\mathcal{O}(N_sKC_s)$, compared with the $\mathcal{O}(N_s^2C_s)$ pairwise term of global self-attention.
This linear scaling supports the high-resolution efficiency gains reported in Section~\ref{sec:segmentation}; full costs are detailed in Appendix~\ref{app:complexity}.

\subsection{Persistent Semantic Memory}
\label{sec:pm}
\textbf{Content- and address-guided retrieval.}
PM retains cross-image semantic experience as $P$ normalized $d_m$-dimensional prototypes for each of $C_{\mathrm{task}}$ categories, forming a bank $M\in\mathbb{R}^{C_{\mathrm{task}}\times P\times d_m}$. Its contents are updated through explicit writes rather than gradients. Both stages share the bank but learn separate readers. 
To retrieve relevant experience, each projected token $z_{s,i}$ produces a normalized content query $q_{s,i}$ and a predicted semantic-address distribution $\pi_{s,i}$. Content similarity identifies related prototypes, while the address softly favors plausible categories:
\begin{equation}
 \ell_{s,i,c,p}
 =
 \frac{
 \langle q_{s,i},m_{c,p}\rangle+
 \lambda_a\log\max(\pi_{s,i,c},\epsilon_a)
 }{\tau_p},
 \label{eq:pm-score}
\end{equation}
where $\lambda_a$ controls the address preference and $\tau_p$ the retrieval temperature. We select the top-$r$ valid prototypes across categories to form $\mathcal{T}_{s,i}$ and aggregate them with softmax weights, as in modern Hopfield networks~\citep{DBLP:conf/iclr/RamsauerSLSWGHA21}:
\begin{equation}
 \alpha_{s,i,c,p}
 =
 \operatorname{softmax}_{(c,p)\in\mathcal{T}_{s,i}}
 (\ell_{s,i,c,p}),
 \qquad
 h_{s,i}
 =
 \sum_{(c,p)\in\mathcal{T}_{s,i}}
 \alpha_{s,i,c,p}m_{c,p}.
 \label{eq:pm-recall}
\end{equation}

\textbf{Persistent-to-working memory binding.}
To combine current and retrieved evidence, we aggregate token-level retrievals into working states using the same normalized write assignment $\bar A_s$:
\begin{equation}
 H_{s,k}=\sum_i\bar A_{s,ik}V_s(h_{s,i}),
 \quad
 \Delta W_s^{\mathrm{PM}}=\sigma(\gamma_s)F_s(H_s),
 \quad
 W_s=W_s^{\mathrm{base}}+\Delta W_s^{\mathrm{PM}}.
 \label{eq:pm-wm-binding}
\end{equation}
$V_s$ maps retrieved vectors to the stage width, $F_s$ projects the pooled content, and $\sigma(\gamma_s)$ is a learned scalar gate. Tokens then read the enriched states through Eq.~\eqref{eq:wm-readback}.

\textbf{Writing and updating PM.}
PM accumulates training evidence from whole images for classification, connected class-specific regions for semantic segmentation, and annotated instance masks for joint detection and instance segmentation. 
For each semantic unit, we pool Stage~3 and Stage~4 features from a separate writer path that excludes current PM retrieval.
Fixed projections map the pooled features to the $d_m$-dimensional prototype space. We normalize each vector, average them, and normalize the result to obtain a descriptor $d$. 

Each descriptor is written to a slot of its annotated class $c$. An unused slot is filled first; otherwise, the most similar prototype is updated by a normalized exponential moving average (EMA):
\begin{equation}
 m_{c,p^*}\leftarrow
 \begin{cases}
 d, & \text{unused slot},\\
 \operatorname{norm}\!\left(
 (1-\eta)m_{c,p^*}+\eta d
 \right),
 & \text{occupied slot},
 \end{cases}
 \label{eq:pm-update}
\end{equation}
where $\eta$ controls how much new evidence is incorporated and $\operatorname{norm}$ denotes $\ell_2$ normalization. We also train the semantic-address heads to predict the annotated categories using auxiliary cross-entropy losses. All PM reads in a training step use the same pre-update bank. New descriptors are committed sequentially only after a successful parameter update. The writer is active only during training; PM is fixed at inference, so the writer introduces no inference-time overhead. Its training-time cost is quantified in Table~\ref{tab:writer_cost} in Appendix~\ref{app:writer_cost}.

\begin{table}[t]
\centering
\caption{
Image classification on the ImageNet-1K. $^\ddagger$ denotes MLP-Mixer parameter counts excluding the classification head. RC, random crop; RRC, random resized crop; IC, Inception-style crop; HF, horizontal flip; RA, RandAugment; MU, Mixup; CM, CutMix; RE, random erasing. $^\star$: stronger augmentation ImageNet-1K pretraining.}
\label{tab:imagenet_classification}
\setlength{\tabcolsep}{2.7pt}
\begin{tabular}{@{}lccclc@{}}
\toprule
Method
& Architecture
& Input
& \#param.
& Data Aug.
& Top-1 acc. \\
\midrule

ResNet-18
& CNN
& $224^{2}$
& 11.7M
& RRC + HF
& 69.8 \\

ResNet-34
& CNN
& $224^{2}$
& 21.8M
& RRC + HF
& 73.3 \\

ResNet-50
& CNN
& $224^{2}$
& 25.6M
& RRC + HF
& 76.1 \\

ResNet-101
& CNN
& $224^{2}$
& 44.5M
& RRC + HF
& 77.4 \\

SE-ResNet-50
& CNN 
& $224^{2}$
& 28.1M
& RRC + HF
& 76.7 \\

ConvNeXt-T
& CNN
& $224^{2}$
& 28.6M
& RRC + HF + RA + MU + CM + RE
& 82.1 \\

\midrule

ViT-B/16
& Transformer
& $384^{2}$
& 86.6M
& RRC + HF
& 77.9 \\

ViT-L/16
& Transformer
& $384^{2}$
& 304.3M
& RRC + HF
& 76.5 \\

ViT-S/16
& Transformer
& $224^{2}$
& 22.1M
& IC + HF + RA + MU
& 78.8 \\

DeiT-S
& Transformer
& $224^{2}$
& 22.1M
& RRC + HF + RA + MU + CM + RE
& 79.8 \\

Swin-T
& Transformer
& $224^{2}$
& 28.3M
& RRC + HF + RA + MU + CM + RE
& 81.2 \\

PVT-T
& Transformer
& $224^{2}$
& 13.2M
& RC + HF + MU + CM + RE
& 75.1 \\

PVT-S
& Transformer
& $224^{2}$
& 24.5M
& RC + HF + MU + CM + RE
& 79.8 \\
\midrule

MLP-Mixer-B/16
& MLP
& $224^{2}$
& 59M$^\ddagger$
& RRC + HF + RA + MU
& 76.4 \\

MLP-Mixer-S/16
& MLP
& $224^{2}$
& 18.5M$^\ddagger$
& RRC + HF + RA + MU + CM + RE
& 74.3 \\

Swin-Mixer-T/D24
& MLP
& $256^{2}$
& 20M
& RRC + HF + RA + MU + CM + RE
& 79.4 \\

\midrule

PoolFormer-S12
& MetaFormer
& $224^{2}$
& 11.9M
& RRC + HF + RA + MU + CM + RE
& 77.2 \\

PoolFormer-S24
& MetaFormer
& $224^{2}$
& 21.4M
& RRC + HF + RA + MU + CM + RE
& 80.3 \\
\midrule

Vim-T
& SSM
& $224^{2}$
& 7M
& RRC + HF + RA + MU + CM + RE
& 76.1 \\

Vim-S
& SSM
& $224^{2}$
& 26M
& RRC + HF + RA + MU + CM + RE
& 80.3 \\

MSVMamba-Micro 
&  SSM 
&  $224^{2}$ 
&  11.9M 
&  RRC + HF + RA + MU + CM + RE 
&  79.8  \\

MambaVision-T 
&  SSM+Trans. 
&  $224^{2}$ 
&  31.8M 
&  RRC + HF + RA + MU + CM + RE 
&  82.3  \\
\midrule

\textbf{DuSAM (ours)}
& \textbf{Memory}
& $224^{2}$
& 20.5M
& RRC + HF + RA
& 77.8 \\

\textbf{DuSAM (ours)$^\star$} 
& \textbf{Memory}
&  $224^{2}$ 
&  20.5M 
&  RRC + HF + RA + MU + CM + RE 
&  79.1  \\

\bottomrule
\end{tabular}
\end{table}

\section{Experiments}
\label{sec:experiments}

\subsection{Image Classification}
\label{sec:classification}
We evaluate DuSAM on ImageNet-1K~\citep{DBLP:conf/cvpr/DengDSLL009}, with training details in Appendix~\ref{app:task-setup}. Table~\ref{tab:imagenet_classification} compares representative backbones under their reported training and evaluation settings. 

DuSAM$^\star$ achieves $79.1$ top-1 accuracy with only $20.5$M parameters, within $0.7$ of DeiT-S and PVT-S while using fewer parameters and no self-attention. It exceeds ResNet-101 by $1.7$ with less than half the parameters and ViT-B/16 by $1.2$ with $76.3\%$ fewer. These comparisons support competitive image-level recognition with a compact memory-native backbone.

\begin{table}[tbp]
\centering
\caption{Semantic segmentation on ADE20K with UPerNet. SS and MS denote single- and multi-scale evaluation; $\dagger$ indicates an additional SimpleFPN is used to produce hierarchical feature maps. $^\star$: stronger augmentation pretraining.}
\label{tab:ade20k_segmentation}
\setlength{\tabcolsep}{4.8pt}
\begin{tabular}{@{}lc*{6}{c}}
\toprule
\multirow{2}{*}{Method}
& \multirow{2}{*}{Architecture}
& \multicolumn{2}{c}{\#param.}
& \multirow{2}{*}{FLOPs }
& \multirow{2}{*}{\shortstack{Train Mem.}}
& \multicolumn{2}{c}{mIoU } \\
\cmidrule(lr){3-4}
\cmidrule(lr){7-8}
&
& Backbone
& Total
&
&
& SS
& MS \\
\midrule

ResNet-18
& CNN
& 11.2M & 41.5M
& 220.29G
& 14.29GiB
& 38.90 & 39.36 \\

ResNet-34
& CNN
& 21.3M & 51.6M
& 229.97G
& 14.77GiB
& 40.08 & 41.66 \\

ResNet-50
& CNN
& 23.5M & 66.5M
& 238.19G
& 22.53GiB
& 41.65 & 42.60 \\

SE-ResNet-50
& CNN 
& 26.0M & 69.0M
& 236.96G
& 23.13GiB
& 41.63 & 41.66 \\

ConvNeXt-T
& CNN 
& 27.8M & 60.2M
& 234.94G
& 18.67GiB
& 45.00 & 45.92 \\

\midrule

ViT-Tiny$^\dagger$
& Transformer
& 5.5M & 37.7M
& 228.31G
& 24.53GiB
& 39.28 & 40.51 \\

ViT-S$^\dagger$
& Transformer
& 21.7M & 53.5M
& 249.35G
& 29.42GiB
& 42.47 & 43.59 \\

DeiT-S$^\dagger$
& Transformer
& 21.7M & 53.5M
& 249.35G
& 29.42GiB
& 43.90 & 44.54 \\

Swin-T
& Transformer
& 27.5M & 59.9M
& 237.19G
& 19.90GiB
& 44.02 & 44.87 \\

PVT-T
&  Transformer
& 12.7M & 43.2M
& 222.63G
& 18.20GiB
& 37.40 & 38.57 \\

PVT-S
& Transformer
& 24.0M & 54.5M
& 233.95G
& 21.99GiB
& 42.46 & 43.79 \\

\midrule

MLP-Mixer-S/16$^\dagger$
& MLP
& 18.0M & 52.6M
& 239.99G
& 36.62GiB
& 33.07 & 33.84 \\

Swin-Mixer-T/D24
& MLP
& 19.6M & 52.0M
& 227.71G
& 21.07GiB
& 40.97 & 42.38\\

\midrule

PoolFormer-S12
& MetaFormer
& 11.4M & 41.9M
& 220.36G
& 16.68GiB
& 39.68 & 40.28 \\

PoolFormer-S24
& MetaFormer
& 21.0M & 51.4M
& 228.66G
& 19.68GiB
& 41.74 & 42.85 \\

\midrule

Vim-T$^\dagger$
& SSM
& 7.1M & 35.7M
& 221.99G
& 16.30GiB
& 36.82 & 39.47 \\

Vim-S$^\dagger$
& SSM
& 25.7M & 57.6M
& 248.33G
& 24.87GiB
& 40.36 & 43.52 \\

MSVMamba-Micro 
&  SSM 
&  11.4M  &  41.7M 
&  218.82G 
&  19.86GiB 
&  45.46  &  45.98  \\

MambaVision-T 
&  SSM+Trans. 
&  31.2M  &  62.5M 
&  322.21G 
&  22.01GiB 
&  46.78  &  47.41  \\

\midrule

\textbf{DuSAM (ours)}
& \textbf{Memory}
& 19.4M & 49.9M
& 226.27G
& 18.98GiB
& 42.90 & 44.12 \\

\textbf{DuSAM (ours)$^\star$} 
& \textbf{Memory}
&  19.4M  &  49.9M 
&  226.27G 
&  18.98GiB 
&  43.33  &  44.50  \\

\bottomrule
\end{tabular}
\end{table}

\begin{figure}[!t]
    \centering
    \includegraphics[width=\linewidth]{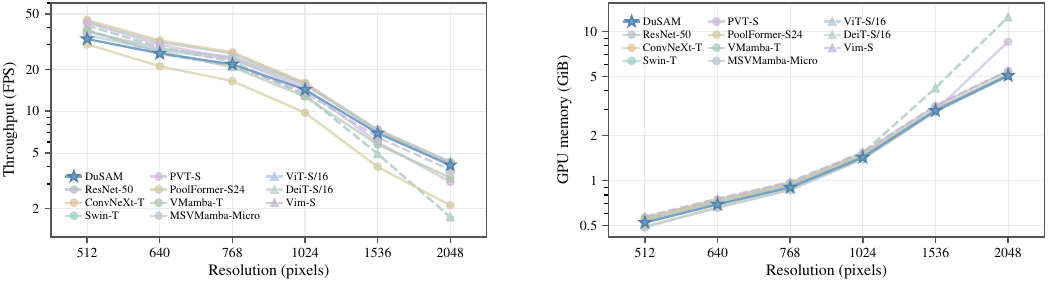}
    \caption{Inference efficiency versus input resolution. \emph{Solid lines}: hierarchical backbones. \emph{Dashed lines}: plain backbones with SimpleFPN. \emph{Left:} Throughput (FPS). \emph{Right:} GPU memory usage (GiB).
    }
    \label{fig:high_resolution}
\end{figure}

\subsection{Semantic Segmentation}
\label{sec:segmentation}
We evaluate DuSAM on ADE20K~\citep{DBLP:conf/cvpr/ZhouZPFB017} with UPerNet~\citep{DBLP:conf/eccv/XiaoLZJS18}. Detailed settings are provided in Appendix~\ref{app:task-setup}. DuSAM$^\star$ achieves $43.33/44.50$ SS/MS mIoU with $19.4$M backbone parameters (Table~\ref{tab:ade20k_segmentation}). It exceeds ResNet-50 by $1.68$ SS mIoU with $25.0\%$ fewer segmentor parameters, and outperforms PVT-S, PoolFormer-S24, and Vim-S by $0.87$, $1.59$, and $2.97$, respectively. Compared with ViT-S, it improves SS/MS mIoU by $0.86/0.91$ while using $35.5\%$ less training memory under common FP16 profiling with 16 images per GPU.

\paragraph{Computational efficiency.} 
We compare DuSAM with ten hierarchical and plain backbones in complete UPerNet segmentors to assess the practical scaling of its linear-cost memory design (Section~\ref{sec:Methodology}). All models use whole-image FP32 inference with batch size 1 on a single NVIDIA H100 80GB GPU. Figure~\ref{fig:high_resolution} shows throughput and peak GPU memory across resolutions; numerical results and FLOPs are reported in Appendix~\ref{app:efficiency} (Table~\ref{tab:inference_efficiency}).

From $512^2$ to $2048^2$, DuSAM's FLOPs increase by $16.0\times$ for $16\times$ more pixels. It retains 12.3\% of its throughput, matching MSVMamba-Micro's retention rate. At $2048^2$, its throughput is within 5.3\% of the highest measured, and its peak GPU memory is only 0.8\% above the lowest. It uses 59.5\% and 40.8\% less GPU memory than DeiT-S/16 and PVT-S, respectively. Kernel profiling suggests that launch overhead in the current implementation limits low-resolution throughput despite DuSAM's low FLOP count, pointing to kernel fusion as a potential optimization (Table~\ref{tab:kernel_profiling}).

\begin{table}[t]
\centering
\caption{Object detection and instance segmentation on COCO val2017 with Mask R-CNN. $\dagger$ indicates an additional SimpleFPN is used. $^\star$: stronger augmentation pretraining.}
\label{tab:coco_detection}
\setlength{\tabcolsep}{6.4pt}
\begin{tabular}{@{}lcc*{6}{c}}
\toprule
\multirow{2}{*}{Method}
& \multirow{2}{*}{Architecture}
& \multirow{2}{*}{\#param.}
& \multicolumn{3}{c}{Bounding box }
& \multicolumn{3}{c}{Instance mask } \\
\cmidrule(lr){4-6}
\cmidrule(lr){7-9}
&
&
& $\mathrm{AP}$
& $\mathrm{AP}_{50}$
& $\mathrm{AP}_{75}$
& $\mathrm{AP}$
& $\mathrm{AP}_{50}$
& $\mathrm{AP}_{75}$ \\
\midrule

ResNet-18 
& CNN
& 31.3M
& 34.0 & 54.0 & 36.8
& 31.2 & 51.0 & 32.9 \\

ResNet-34 
& CNN
& 41.2M
& 37.2 & 57.5 & 40.5
& 33.7 & 54.3 & 36.1 \\

ResNet-50
& CNN
& 44.2M
& 38.0 & 58.6 & 41.4
& 34.4 & 55.1 & 36.7 \\

ResNet-101
& CNN
& 63.2M
& 40.0 & 60.5 & 44.0
& 36.1 & 57.5 & 38.6 \\

SE-ResNet-50
& CNN
& 46.7M
& 38.7 & 60.9 & 42.1
& 35.4 & 57.4 & 37.8 \\

ConvNeXt-T 
& CNN
& 48.1M
& 43.6 & 65.5 & 48.1
& 40.3 & 62.9 & 43.4 \\

\midrule

ViT-T $^\dagger$
& Transformer
& 27.9M
& 35.2 & 57.5 & 37.2
& 33.3 & 54.5 & 34.8 \\

ViT-S $^\dagger$
& Transformer
& 43.7M
& 39.6 & 62.5 & 42.5
& 36.7 & 59.1 & 38.7 \\

DeiT-S $^\dagger$
& Transformer
& 43.7M
& 40.4 & 63.3 & 43.5
& 37.1 & 59.9 & 39.0 \\

Swin-T 
& Transformer
& 47.8M
& 42.6 & 65.0 & 46.5
& 39.2 & 62.1 & 42.3 \\

PVT-T
& Transformer
& 32.9M
& 36.7 & 59.2 & 39.3
& 35.1 & 56.7 & 37.3 \\

PVT-S
& Transformer
& 44.1M
& 40.4 & 62.9 & 43.8
& 37.8 & 60.1 & 40.3 \\

\midrule

MLP-Mixer-S/16 $^\dagger$
& MLP
& 41.6M
& 28.3 & 49.0 & 28.9
& 27.5 & 46.2 & 28.6 \\

Swin-Mixer-T/D24 
& MLP
& 39.9M
& 39.9 & 62.4 & 43.2
& 36.9 & 59.2 & 39.2 \\

\midrule

PoolFormer-S12
& MetaFormer
& 31.6M
& 37.3 & 59.0 & 40.1
& 34.6 & 55.8 & 36.9 \\

PoolFormer-S24
& MetaFormer
& 41.0M
& 40.1 & 62.2 & 43.4
& 37.0 & 59.1 & 39.6 \\

\midrule

Vim-T $^\dagger$
& SSM
& 27.2M
& 34.2 & 54.8 & 36.4
& 31.4 & 51.7 & 32.7 \\

Vim-S $^\dagger$
& SSM
& 46.3M
& 39.1 & 60.3 & 42.0
& 35.3 & 56.9 & 37.0 \\

 MSVMamba-Micro 
&  SSM 
&  32.0M 
&  43.8  &  65.8  &  48.1 
&  39.9  &  62.7  &  42.8  \\

 MambaVision-T 
&  SSM+Trans. 
&  51.4M 
&  45.2  &  68.3  &  49.2 
&  41.5  &  65.2  &  44.6  \\

\midrule

\textbf{DuSAM (ours)}
& \textbf{Memory}
& 39.5M
& 41.9 & 64.1 & 45.9
& 38.1 & 60.9 & 40.6 \\

\textbf{DuSAM (ours)$^\star$} 
& \textbf{Memory} 
&  39.5M 
&  42.6  &  65.1  &  46.3 
&  38.9  &  62.0  &  41.7  \\

\bottomrule
\end{tabular}
\end{table}

\begin{table}[t]
\centering
\caption{Component ablations on ADE20K. Classes are divided into four approximately equal-sized groups by annotated training-pixel count, from rare (Lowest) to frequent (Highest).}
\label{tab:memory_ablation_frequency}
\setlength{\tabcolsep}{3.0pt}
\begin{tabular}{@{}lccc*{5}{c}}

\toprule
\multirow{2}{*}{Method}
& \multirow{2}{*}{\#param.}
& \multirow{2}{*}{PM state}
& \multirow{2}{*}{Train Mem.}
& \multicolumn{5}{c}{SS mIoU } \\

\cmidrule(lr){5-9}

& & & &
Overall
& Lowest
& Low
& High
& Highest \\

\midrule

\textbf{Standard}
& 19.43M & 0.77MiB & 18.98GiB
& $\mathbf{42.90}$
& $\mathbf{37.75}$
& $\mathbf{34.81}$
& $\mathbf{42.96}$
& $\mathbf{56.43}$ \\

\midrule

w/o WM+PM
& 17.18M & 0.00MiB & 18.18GiB
& $40.45$
& $33.96$
& $30.62$
& $42.00$
& $55.64$ \\

w/o WM transport
& 19.43M & 0.77MiB & 18.99GiB
& $42.48$
& $36.50$
& $34.29$
& $43.23$
& $56.29$ \\

w/o PM
& 18.16M & 0.00MiB & 18.58GiB
& $41.75$
& $35.87$
& $31.96$
& $43.28$
& $56.30$ \\

w/o PM, keep address
& 18.53M & 0.00MiB & 18.65GiB
& $41.55$
& $35.20$
& $31.91$
& $43.38$
& $56.13$ \\

Separate PM ($2{+}2$)
& 19.43M & 0.77MiB & 18.99GiB
& $42.39$
& $37.30$
& $33.49$
& $43.00$
& $56.14$ \\

Separate PM ($4{+}4$)
& 19.43M & 1.53MiB & 18.99GiB
& $42.37$
& $37.71$
& $32.91$
& $42.93$
& $56.29$ \\

\bottomrule
\end{tabular}
\end{table}

\begin{table}[t]
\centering
\setlength{\tabcolsep}{4pt}
\renewcommand{\arraystretch}{1.06}
\caption{Comparison with existing mechanisms on ADE20K. We replace either the PM injection route or the WM write mechanism in DuSAM, keeping the remaining components unchanged. $^{\clubsuit}$ indicates that DuSAM's WM write-side parameters are reinitialized. Class groups follow Table~\ref{tab:memory_ablation_frequency}.}
\label{tab:mechanism_comparison}

\begin{tabular*}{\linewidth}
{@{\extracolsep{\fill}}lrrrrrr@{}}
\toprule
\multicolumn{1}{l}{\multirow{2}{*}[-0.6ex]{Variant}}
& \multicolumn{1}{c}{\multirow{2}{*}[-0.6ex]{\#param.}}
& \multicolumn{5}{c}{SS mIoU} \\
\cmidrule(l){3-7}
& & Overall & Lowest & Low & High & Highest \\
\midrule

\multicolumn{7}{@{}l}{
    \textit{\textbf{Persistent-memory injection route}}
} \\
\quad PM-to-WM binding (ours)
& 19.43M
& 42.90 & 37.75 & 34.81 & 42.96 & 56.43 \\
\quad PM-to-token fusion
& 19.43M
& 42.03 & 37.45 & 31.59 & 43.08 & 56.41 \\

\midrule

\multicolumn{7}{@{}l}{
    \textit{\textbf{Working-memory write mechanism}}
} \\
\quad Spatially anchored writing (ours)$^{\clubsuit}$
& 19.43M
& 42.78 & 36.31 & 35.54 & 43.12 & 56.51 \\
\quad Prior-guided bottleneck attention
& 19.43M
& 42.28 & 36.88 & 33.68 & 42.92 & 56.01 \\
\quad Pooled-query cross-attention
& 19.41M
& 41.98 & 33.96 & 33.29 & 43.97 & 57.16 \\

\bottomrule
\end{tabular*}
\end{table}

\subsection{Object Detection and Instance Segmentation}
\label{sec:detection}
We evaluate DuSAM on COCO 2017~\citep{DBLP:conf/eccv/LinMBHPRDZ14} using Mask R-CNN~\citep{DBLP:conf/iccv/HeGDG17} with FPN~\citep{DBLP:conf/cvpr/LinDGHHB17}; detailed settings are provided in Appendix~\ref{app:task-setup}. DuSAM$^\star$ achieves $42.6$ box AP and $38.9$ mask AP with $39.5$M detector parameters (Table~\ref{tab:coco_detection}). It matches Swin-T in box AP with $17\%$ fewer parameters and exceeds ResNet-101 by $2.6/2.8$ box/mask AP with $37.5\%$ fewer parameters. These gains extend across all three object-size groups (Appendix~\ref{app:coco_size}). It outperforms DeiT-S, Swin-Mixer-T/D24, and Vim-S by $2.2/1.8$, $2.7/2.0$, and $3.5/3.6$, respectively, while using fewer parameters in each comparison. The gains over DeiT-S also extend to the stricter AP$_{75}$ metric, reaching $2.8$ for boxes and $2.7$ for masks.

\subsection{Ablations and Memory Analysis}
\label{sec:memory_analysis}
We analyze how working and persistent memory structure visual representations on ADE20K.

\textbf{Memory configuration.}
Table~\ref{tab:memory_ablation_frequency} shows that WM improves over the local-only model, while PM provides its largest gains in the two rarer class groups. Retaining the auxiliary category-prediction heads and losses while removing PM (\emph{w/o PM, keep address}) still lowers accuracy, so additional supervision alone cannot explain PM's benefit. Removing cross-stage transport also reduces accuracy. Separate banks with $2{+}2$ prototypes per class across stages match total capacity, whereas $4{+}4$ doubles it. Both comparisons favor a shared semantic store across stages. Appendix~\ref{app:frozen_transfer} further examines frozen-feature transfer under different PM settings.

\textbf{Mechanism comparison.}
As shown in Table~\ref{tab:mechanism_comparison}, we compare alternatives from prior work within DuSAM under the same training and evaluation settings. For PM injection, we replace our PM-to-WM route with PM-to-token fusion inspired by MCIBI~\citep{DBLP:conf/iccv/JinGY0WWS21}. The latter retains PM retrieval and WM while injecting the retrieved content into dense tokens. Overall SS mIoU decreases from 42.90 to 42.03, with the largest drop in the Low group (3.22 points). This suggests that integrating persistent semantic evidence with image-local context in WM improves its use in dense prediction. For WM writing, we replace our spatially anchored Sinkhorn assignment with learned-prior top-$b$ bottleneck attention inspired by AiT~\citep{DBLP:conf/cvpr/SunOW0K25} or pooled-query cross-attention inspired by LookupViT~\citep{DBLP:conf/eccv/KonerJJTP24}. WM capacity and the original readback remain unchanged. For a fair comparison, we also reinitialize the WM write-side parameters of our reference, as the replacement mechanisms introduce randomly initialized parameters. Our reference achieves 42.78 SS mIoU, compared with 42.28 and 41.98 for the two alternatives, respectively. At fixed WM capacity, the performance gap points to the importance of how image evidence is organized into working states.

\begin{figure}[!t]
    \centering
    \includegraphics[width=\linewidth]{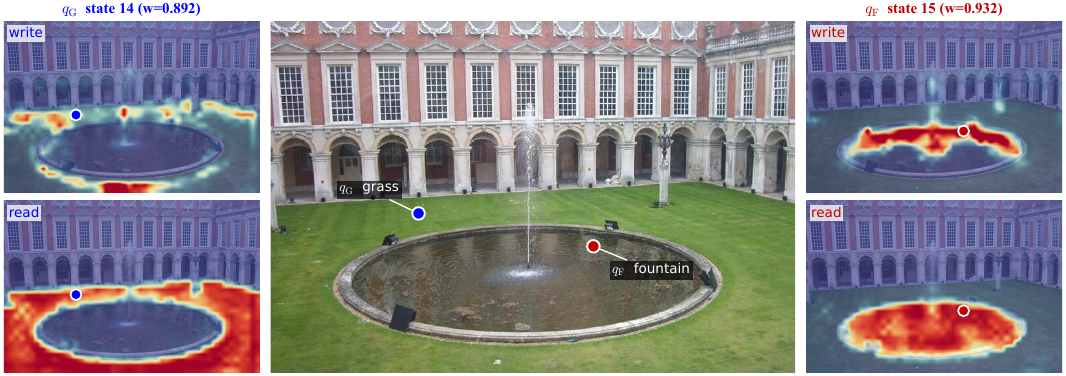}
    \caption{Visualization of WM routing for two queries.}
    \label{fig:wm_routing}
\end{figure}

\begin{figure}[t]
    \centering
    \includegraphics[width=\linewidth]{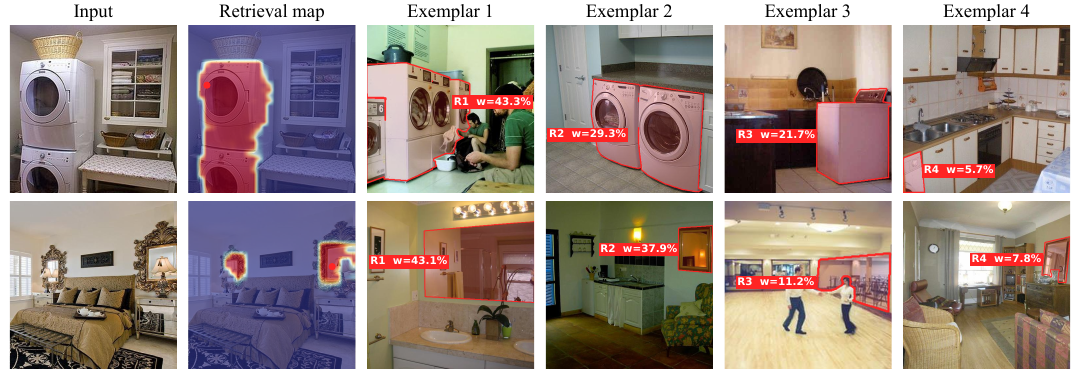}
    \caption{Visualization of PM retrieval. Maps show target-class retrieval weights; exemplars are nearest training regions of the top-four prototypes at the marked query. $w$ are not renormalized.}
    \label{fig:pm_retrieval}
\end{figure}

\begin{wraptable}[17]{r}{0.46\linewidth}
\vspace{-13pt}
\centering
\caption{Inference-time memory interventions on ADE20K. Disagr.: pixel-label disagreement with Standard.}
\label{tab:memory_interventions}
\setlength{\tabcolsep}{2.5pt}
\begin{tabular}{@{}lrrr}
\toprule
Mode
& mIoU 
& $\Delta$mIoU
& \shortstack{Disagr.} \\
\midrule

\textbf{Standard}
& $\mathbf{42.90}$
& $\mathbf{0.00}$
& $\mathbf{0.00}$\\

\midrule

WM\_off
& 11.01
& $-31.89$
& 57.02 \\

WM\_shuffle
& 20.01
& $-22.89$
& 34.02 \\

WM\_mean\_state
& 37.10
& $-5.80$
& 11.93 \\

WM\_rank1\_state
& 39.07
& $-3.83$
& 9.98 \\

WM\_transport\_off
& 38.19
& $-4.71$
& 11.54 \\

\midrule

PM\_off
& 42.01
& $-0.89$
& 5.65 \\

PM\_shuffle
& 41.50
& $-1.40$
& 6.93 \\

PM\_wrong\_class
& 41.94
& $-0.96$
& 6.62 \\

\bottomrule
\end{tabular}
\end{wraptable}

\textbf{WM spatial routing.} To illustrate how compact workspace is used, we visualize two queries in Fig.~\ref{fig:wm_routing}, using Stage~3 as an example. For each query, we show highest-read-weight state's current write sources and spatial read weights. The grass and fountain queries access different states (states~14 and~15 with weights $0.892$ and $0.932$), whose maps emphasize distinct scene regions, illustrating how compact workspace provides different locations with relevant spatial context. With parameters fixed, disabling WM or shuffling its channels causes large accuracy drops. Mean-state and rank-1 replacements reduce mIoU by $5.80$ and $3.83$, while removing transport costs $4.71$ (Table~\ref{tab:memory_interventions}). The predictor therefore exploits both distinctions among states and context propagated across stages. Appendix~\ref{app:wm_cross_stage} tracks how WM writes and reads evolve across stages, while Appendix~\ref{app:memory_capacity} further analyzes WM capacity and utilization.

\textbf{PM semantic retrieval.}
To visualize this semantic access, Fig.~\ref{fig:pm_retrieval} shows two examples at Stage~3. Each map sums retrieval weights for the target class across image locations; the accompanying exemplars are nearest training regions of the marked query's top-four prototypes. Together, they show where class-specific memory is used and the visual patterns associated with its retrieved contents. Additional examples are provided in Appendix~\ref{app:pm_retrieval_examples}. PM capacity, utilization, and sensitivity to $\eta$, $r$, and $\tau_p$ are further analyzed in Appendix~\ref{app:memory_capacity}.

\textbf{Class-targeted memory updates.} 
To test whether stored experience can be revised selectively, we remove each class's PM entries and restore them by applying the training-time writer to five randomly sampled annotated regions of that class, keeping network parameters fixed. Across all 150 ADE20K classes, mean target-class IoU falls from 42.90 to 42.27 after removal and recovers to 42.76 after restoration, i.e., 78\% of the lost IoU, while the mean absolute IoU change of other classes is only 0.02 pp. Removal also enables class-level diagnosis of PM (Table~\ref{tab:negatives} in Appendix~\ref{app:pm_restoration_examples}).

\begin{figure}[t!]
    \centering
    \includegraphics[width=\linewidth]{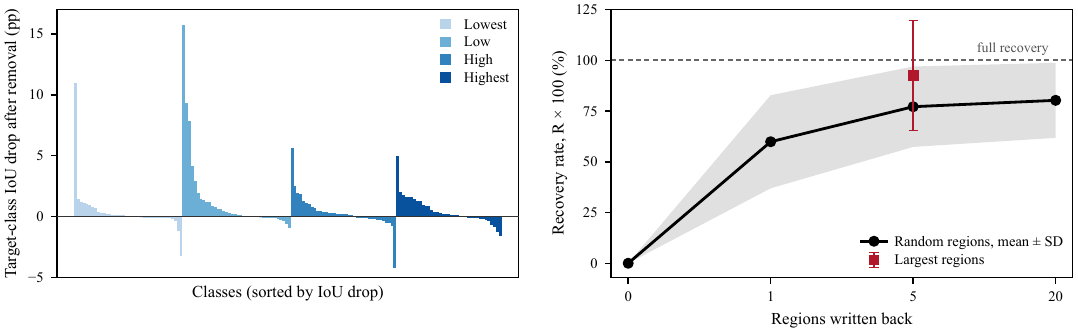}
    \caption{
    Class-targeted PM removal and restoration on ADE20K. \emph{Left:} target-class IoU drops, sorted within annotated-pixel frequency groups. \emph{Right:} recovery rates (mean $\pm$ SD across classes).
    }
    \label{fig:pm_edit_stats}
\end{figure}

\begin{figure}[t]
    \centering
    \includegraphics[width=\linewidth]{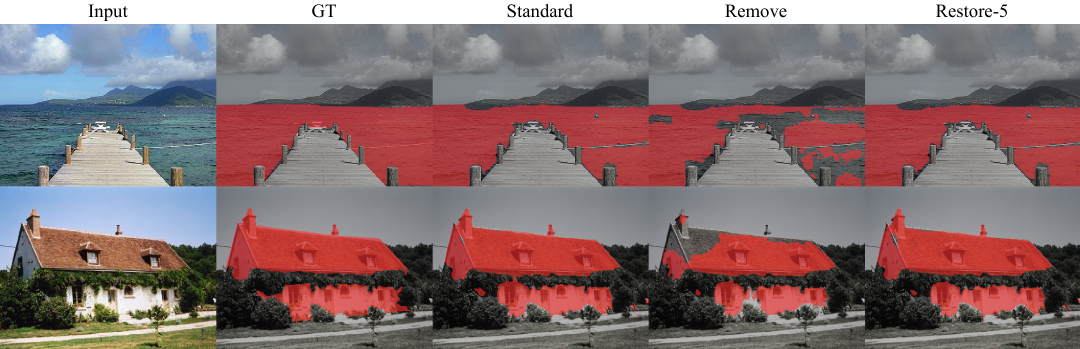}
    \caption{Class-targeted PM removal and restoration with fixed weights. Restore-5 uses five annotated training regions. Red marks the target class on backgrounds desaturated for clarity.
}
    \label{fig:pm_restoration}
\end{figure}

Reliance on PM varies across classes (Fig.~\ref{fig:pm_edit_stats} \emph{Left}). Since removal leaves network parameters intact, some classes change little. This is expected: PM is designed to complement learned parameters and image-local WM, and matters most for classes these capture less well, such as those with fewer annotated training pixels. We therefore examine restoration on classes whose IoU drops by at least 1 pp, where recovery can be measured reliably (Fig.~\ref{fig:pm_edit_stats} \emph{Right}). Their mean drop is 3.1 and 4.7 pp in the Lowest and Low groups versus 2.2 and 2.0 pp in the High and Highest groups, consistent with the group-wise PM benefits in Table~\ref{tab:memory_ablation_frequency}. Their mean IoU falls from 43.00 to 39.91 and recovers to 42.24, a mean per-class recovery of 77\% (67\%--92\% across groups) that is stable across three random draws (mean within-class SD of 0.41 pp), while other classes change by 0.11 pp. One region already recovers 60\%, and the five largest recover 93\%. Fig.~\ref{fig:pm_restoration} and Appendix~\ref{app:pm_restoration_examples} show qualitative examples on ADE20K and COCO. DuSAM thus supports targeted memory updates from a few annotated regions without changing network parameters.

\section{Conclusion}
\label{sec:conclusion}
We introduced DuSAM, a memory-native visual backbone organized around image-local working memory and persistent semantic memory rather than self-attention or state-space blocks. It achieves competitive performance on classification, segmentation, and detection, scales efficiently to high resolution, and stores semantic evidence that can be inspected and replenished without weight updates. These results establish explicit memory as a viable organizing principle for visual backbones.

\subsubsection*{Acknowledgments}
This research was funded in part by the Austrian Science Fund (FWF) 10.55776/COE12 and the AXA Research Fund. 

\bibliography{iclr2027_conference}
\bibliographystyle{iclr2027_conference}

\newpage
\appendix
\section{Appendix}

\subsection{Implementation Details}
\label{app:implementation}

\subsubsection{Model Details}
\label{app:model-config}
The four stages use widths $(64,128,320,512)$ and local depths $(3,3,8,3)$. Local embedding comprises two overlapping $3\times3$, stride-$2$ projections with output widths $32$ and $64$, each followed by batch normalization and ReLU. Spatial reduction uses a single $3\times3$, stride-$2$ projection. Each local block combines a pre-normalized pointwise residual with a pre-normalized channel-expansion branch using expansion ratio $4$, GELU, and channel-wise $3\times3$ mixing. WM uses $K=21$ states anchored at the centers of $1\times1$, $2\times2$, and $4\times4$ grids, with $\tau_w=0.1$, $\lambda_{\mathrm{sp}}=8$, and seven log-Sinkhorn iterations. PM uses $P=4$ prototypes per class with $d_m=320$, $r=8$, $\tau_p=0.1$, $\lambda_a=0.25$, and $\eta=0.05$. Following layer normalization, the query head projects features to $d_m$ dimensions; the address head uses two linear layers with hidden width $d_m$ and $C_{\mathrm{task}}$ outputs. Both memory gates are initialized to output $0.05$. Invalid PM slots are excluded from retrieval, and an empty bank leaves the working state unchanged. The semantic-address losses at Stages~3 and~4 each have weight $0.1$.

\subsubsection{Training and Evaluation}
\label{app:task-setup}

\paragraph{Image classification.}
We train DuSAM from scratch on ImageNet-1K~\citep{DBLP:conf/cvpr/DengDSLL009}, using approximately $1.28$ million training images and $50{,}000$ validation images across $1000$ categories. Training uses $224\times224$ inputs, AdamW~\citep{DBLP:conf/iclr/LoshchilovH19}, a total batch size of $128$, and an initial learning rate of $10^{-3}$ with cosine decay over $300$ epochs. Augmentation consists of random resized cropping, horizontal flipping, and RandAugment with two operations and magnitude nine. The classification loss is cross-entropy with label smoothing $0.1$. DuSAM$^\star$ additionally uses Mixup with $\alpha=0.8$, CutMix with $\alpha=1.0$, random erasing with probability 0.25, and stochastic depth with rate 0.1. All other settings remain unchanged. Mixing is applied to 80\% of batches; the classifier and semantic-address heads use the corresponding soft targets, and PM is written from the unmixed batches so that each prototype accumulates evidence from a single class. We report top-1 accuracy on the validation set. 

\paragraph{Semantic segmentation.} 
We evaluate on ADE20K~\citep{DBLP:conf/cvpr/ZhouZPFB017}, with $20{,}210$ training and $2000$ validation images from $150$ categories. UPerNet \citep{DBLP:conf/eccv/XiaoLZJS18} uses ImageNet-1K-pretrained weights to initialize the transferable backbone parameters. Training runs for $160$k iterations with $512\times512$ crops, global batch size $16$, AdamW, learning rate $6\times10^{-5}$, and weight decay $0.01$. A $1500$-iteration linear warmup precedes polynomial decay with power one. The decoder and auxiliary cross-entropy losses have weights $1.0$ and $0.4$, respectively. We report single-scale (SS) and multi-scale (MS) mIoU; MS combines scales $\{0.5,0.75,1.0,1.25,1.5,1.75\}$ with horizontal flipping. FLOPs in Table~\ref{tab:ade20k_segmentation} are reported for a single $512\times512$ input using MMEngine~\citep{mmengine2022}. GPU memory is reported as peak allocated memory.

\paragraph{Object detection and instance segmentation.} We use Mask R-CNN~\citep{DBLP:conf/iccv/HeGDG17} with FPN~\citep{DBLP:conf/cvpr/LinDGHHB17} on COCO 2017~\citep{DBLP:conf/eccv/LinMBHPRDZ14} with $80$ foreground categories. Training retains $117{,}266$ train2017 images after annotation and minimum-size filtering; evaluation uses all $5000$ val2017 images. The backbone is initialized from the ImageNet-1K-pretrained DuSAM weights, with randomly initialized FPN and detection heads. We train for $12$ epochs with a total batch size of $16$, using AdamW with learning rate $1.5\times10^{-4}$ and weight decay $0.05$. A $500$-update linear warmup starts at $10^{-3}$ times the base learning rate, followed by tenfold decays at epochs $8$ and $11$. Images retain their aspect ratio, with the shorter side resized to $800$ pixels and the longer side capped at $1333$ pixels; random horizontal flipping with probability $0.5$ is the only additional augmentation. Evaluation uses the same resizing rule and single-scale inference without test-time augmentation. We report box and mask AP, AP$_{50}$, AP$_{75}$, and size-specific $\mathrm{AP}_{s}$, $\mathrm{AP}_{m}$, $\mathrm{AP}_{l}$.

\paragraph{Parameter transfer and PM initialization.}
For ADE20K and COCO, we initialize DuSAM from ImageNet-1K pretrained weights, retaining compatible parameters for WM construction, cross-stage transport, readback, PM content querying, and PM-to-WM binding. The task-independent layers of the semantic-address heads are transferred, while their class-specific output layers are reinitialized for the downstream label vocabulary (150 classes for ADE20K and 80 for COCO). The PM prototype bank is reinitialized and populated with downstream training regions or instances rather than ImageNet prototypes. The PM writer follows the same descriptor-construction and EMA update procedure, using target-task annotations to accumulate new semantic evidence. The task-specific segmentation and detection heads are initialized separately.

\paragraph{Baselines.}
\label{app:baseline}
We compare DuSAM with representative CNN, Transformer, MLP, MetaFormer, and SSM backbones across classification, semantic segmentation, and object detection tasks. 
The baselines include ResNet variants \citep{DBLP:conf/cvpr/HeZRS16}, SE-ResNet-50 \citep{DBLP:conf/cvpr/HuSS18}, ConvNeXt-T \citep{DBLP:conf/cvpr/0003MWFDX22}, ViT-T/S \citep{DBLP:conf/iclr/DosovitskiyB0WZ21}, DeiT-S \citep{DBLP:conf/icml/TouvronCDMSJ21}, Swin-T \citep{DBLP:conf/iccv/LiuL00W0LG21}, PVT-T/S \citep{DBLP:conf/iccv/WangX0FSLL0021}, MLP-Mixer-S/16  \citep{DBLP:conf/nips/TolstikhinHKBZU21}, Swin-Mixer-T/D24 \citep{DBLP:conf/iccv/LiuL00W0LG21}, PoolFormer-S12/S24 \citep{DBLP:conf/cvpr/YuLZSZWFY22}, Vim-T/S \citep{DBLP:conf/icml/ZhuL0W0W24}, VMamba-T \citep{DBLP:conf/nips/LiuTZYX0YJ024}, MambaVision-T~\citep{DBLP:conf/cvpr/HatamizadehK25}, and MSVMamba-Micro~\citep{DBLP:conf/nips/ShiDX24}. 
All baselines trained by us are initialized from publicly released ImageNet-1K checkpoints and follow the same segmentation and detection settings as DuSAM described above, with architecture-specific optimizer hyperparameters. Table~\ref{tab:imagenet_classification} lists published ImageNet-1K results with their augmentation recipes.

\begin{table}[p]
\centering
\caption{UPerNet inference efficiency across input resolutions. Throughput and GPU memory are the median and maximum over three measurement runs, respectively. Retained is the percentage of throughput at $512^2$ maintained at $2048^2$. Growth indicates the corresponding increase factor for peak memory or FLOPs. The pixel count increases $16\times$. $\dagger$ indicates an additional SimpleFPN is used to produce hierarchical feature maps.}
\label{tab:inference_efficiency}

\begin{tabular*}{\linewidth}{@{\extracolsep{\fill}}l*{7}{c}@{}}

\toprule[0.9pt]
\multicolumn{8}{c}{\textbf{Throughput (FPS)} $\uparrow$} \\
\cmidrule(lr){1-8}

\multirow{2}{*}{Method}
& \multicolumn{6}{c}{Input resolution}
& \multirow{2}{*}{
    \begin{tabular}{@{}c@{}}
    Retained\\
    ($\%$) $\uparrow$
    \end{tabular}
} \\

\cmidrule(lr){2-7}
& $512^2$ & $640^2$ & $768^2$
& $1024^2$ & $1536^2$ & $2048^2$ & \\

\midrule
ResNet-50
& 37.569 & 28.267 & 24.543 & 15.780 & 7.185 & 4.207 & 11.20 \\
ViT-S/16$^\dagger$
& 41.077 & 28.848 & 22.788 & 13.069 & 4.944 & 1.736 & 4.23 \\
DeiT-S/16$^\dagger$
& 41.600 & 28.833 & 22.985 & 13.144 & 4.945 & 1.745 & 4.19 \\
Swin-T
& 43.829 & 31.355 & 25.986 & 15.800 & 7.341 & 4.273 & 9.75 \\
ConvNeXt-T
& 45.369 & 32.245 & 26.523 & 16.015 & 7.322 & 4.271 & 9.41 \\
PVT-S
& 43.361 & 30.078 & 23.866 & 14.566 & 5.982 & 3.102 & 7.15 \\
PoolFormer-S24
& 30.282 & 21.036 & 16.474 & 9.760 & 3.984 & 2.103 & 6.94 \\
Vim-S$^\dagger$
& 38.590 & 25.886 & 21.148 & 13.643 & 6.483 & 3.796 & 9.84 \\
VMamba-T
& 37.895 & 26.776 & 20.792 & 12.734 & 5.766 & 3.326 & 8.78 \\
MSVMamba-Micro
& 35.211 & 27.367 & 23.636 & 15.335 & 7.399 & 4.338 & 12.32 \\

\cmidrule(lr){1-8}
\textbf{DuSAM (ours)}
& 33.490 & 25.392 & 21.643 & 14.116 & 6.931 & 4.110 & 12.27 \\

\specialrule{0.9pt}{0.8ex}{0.8ex}
\multicolumn{8}{c}{\textbf{GPU memory (GiB)} $\downarrow$} \\
\cmidrule(lr){1-8}

\multirow{2}{*}{Method}
& \multicolumn{6}{c}{Input resolution}
& \multirow{2}{*}{
    \begin{tabular}{@{}c@{}}
    Growth\\
    ($\times$) $\downarrow$
    \end{tabular}
} \\

\cmidrule(lr){2-7}
& $512^2$ & $640^2$ & $768^2$
& $1024^2$ & $1536^2$ & $2048^2$ & \\

\midrule
ResNet-50
& 0.565 & 0.747 & 0.968 & 1.536 & 3.154 & 5.419 & 9.59 \\
ViT-S/16$^\dagger$
& 0.554 & 0.736 & 0.960 & 1.531 & 4.174 & 12.487 & 22.54 \\
DeiT-S/16$^\dagger$
& 0.554 & 0.736 & 0.960 & 1.531 & 4.174 & 12.487 & 22.54 \\
Swin-T
& 0.560 & 0.730 & 0.941 & 1.474 & 3.002 & 5.139 & 9.18 \\
ConvNeXt-T
& 0.561 & 0.733 & 0.945 & 1.477 & 3.005 & 5.142 & 9.17 \\
PVT-S
& 0.534 & 0.704 & 0.914 & 1.441 & 2.951 & 8.532 & 15.98 \\
PoolFormer-S24
& 0.492 & 0.663 & 0.871 & 1.398 & 2.909 & 5.022 & 10.21 \\
Vim-S$^\dagger$
& 0.575 & 0.757 & 0.981 & 1.551 & 3.180 & 5.457 & 9.49 \\
VMamba-T
& 0.540 & 0.710 & 0.922 & 1.454 & 2.982 & 5.119 & 9.48 \\
MSVMamba-Micro
& 0.487 & 0.657 & 0.864 & 1.392 & 2.902 & 5.013 & 10.29 \\

\cmidrule(lr){1-8}
\textbf{DuSAM (ours)}
& 0.523 & 0.692 & 0.900 & 1.430 & 2.939 & 5.053 & 9.66 \\

\specialrule{0.9pt}{0.8ex}{0.8ex}
\multicolumn{8}{c}{\textbf{FLOPs (G)} $\downarrow$} \\
\cmidrule(lr){1-8}

\multirow{2}{*}{Method}
& \multicolumn{6}{c}{Input resolution}
& \multirow{2}{*}{
    \begin{tabular}{@{}c@{}}
    Growth\\
    ($\times$) $\downarrow$
    \end{tabular}
} \\

\cmidrule(lr){2-7}
& $512^2$ & $640^2$ & $768^2$
& $1024^2$ & $1536^2$ & $2048^2$ & \\

\midrule
ResNet-50
& 238.19 & 372.15 & 535.87 & 952.62 & 2143.32 & 3810.31 & 16.00 \\
ViT-S/16$^\dagger$
& 249.35 & 398.09 & 588.18 & 1113.28 & 2939.70 & 6308.44 & 25.30 \\
DeiT-S/16$^\dagger$
& 249.35 & 398.09 & 588.18 & 1113.28 & 2939.70 & 6308.44 & 25.30 \\
Swin-T
& 237.18 & 368.46 & 530.69 & 945.08 & 2118.61 & 3771.84 & 15.90 \\
ConvNeXt-T
& 234.94 & 367.08 & 528.58 & 939.69 & 2114.28 & 3758.70 & 16.00 \\
PVT-S
& 233.95 & 368.96 & 537.32 & 982.48 & 2385.71 & 4677.20 & 19.99 \\
PoolFormer-S24
& 228.66 & 357.28 & 514.48 & 914.61 & 2057.87 & 3658.42 & 16.00 \\
Vim-S$^\dagger$
& 248.33 & 388.01 & 558.73 & 993.28 & 2234.87 & 3973.10 & 16.00 \\
VMamba-T
& 240.98 & 376.52 & 542.18 & 963.85 & 2168.65 & 3855.36 & 16.00 \\
MSVMamba-Micro
& 218.79 & 341.85 & 492.26 & 875.11 & 1968.99 & 3500.41 & 16.00 \\

\cmidrule(lr){1-8}
\textbf{DuSAM (ours)}
& 226.27 & 353.54 & 509.09 & 905.04 & 2036.32 & 3620.10 & 16.00 \\

\bottomrule[0.9pt]
\end{tabular*}
\end{table}


\begin{table}[t]
\centering
\caption{Kernel-level profiling of complete UPerNet segmentors with DuSAM and ConvNeXt-T backbones. Memory-op shares report the percentages of kernel launches and GPU time attributable to WM and PM operations. GPU busy fraction is the percentage of forward wall-clock time during which at least one kernel is executing.}
\label{tab:kernel_profiling}

\setlength{\tabcolsep}{2.5pt}
\begin{tabular}{@{}lccc@{}}
\toprule

\textbf{Metric}
& \textbf{Resolution}
& \textbf{ConvNeXt-T} 
& \textbf{DuSAM (ours)}
\\

\midrule

\multirow{2}{*}{Kernel launches per forward}
& $512^2$ & 259  & 1,281 \\
& $2048^2$ & 263 & 1,265 \\

\midrule

\multirow{2}{*}{Memory-op shares, launches / GPU time}
& $512^2$  & - / - & 66\% / 3\% \\
& $2048^2$ & - / - & 66\% / 2\% \\

\midrule

\multirow{2}{*}{GPU busy fraction}
& $512^2$  & 96\% & 56\% \\
& $2048^2$ & 99.6\% & 95\% \\

\bottomrule
\end{tabular}
\end{table}

\subsection{Computational Complexity and Efficiency}

\paragraph{Computational Complexity.}
\label{app:complexity}
We analyze how DuSAM's memory computation scales with image resolution. For one image at stage $s$, let $N_s=H_sW_s$ denote the number of tokens, $C_s$ the feature width, and $K$ the number of working states. PM contains at most $M_0=C_{\mathrm{task}}P$ prototypes of dimension $d_m$, with $P$ prototypes for each of $C_{\mathrm{task}}$ categories.

WM matches tokens to working states and aggregates their contents for writing and reading, requiring $\mathcal{O}(N_sKC_s)$ operations. Its $J$ balancing passes add $\mathcal{O}(JN_sK)$. Token and readback projections cost $\mathcal{O}(N_sC_s^2)$, while cross-stage state transport adds $\mathcal{O}(KC_{s-1}C_s)$ when present. 

PM scores candidate prototypes at cost $\mathcal{O}(N_sM_0d_m)$ and aggregates the $r$ selected prototypes at cost $\mathcal{O}(N_srd_m)$. Query and address mappings, recall projection, and binding to WM together add $\mathcal{O}(N_sC_sd_m+N_sd_mC_{\mathrm{task}} +N_sKC_s+KC_s^2)$. Top-$r$ selection is performed per token over at most $M_0$ candidates; it reduces value aggregation but not candidate scoring.

For fixed widths, capacities, and iteration counts, these operations scale linearly with $N_s$, in contrast to the $\mathcal{O}(N_s^2C_s)$ pairwise term of global self-attention. This analysis covers memory computation and its projections; local processing, the training-only writer (quantified in Table~\ref{tab:writer_cost}), and task heads contribute additional costs. State storage is independent of image resolution, whereas token features and materialized matching tensors grow with token count.

\paragraph{Inference Efficiency.}
\label{app:efficiency}
Table~\ref{tab:inference_efficiency} extends the comparison in Figure~\ref{fig:high_resolution} with measurements of all eleven segmentors at six resolutions, including forward FLOP counts. DuSAM maintains low arithmetic cost across all six resolutions, using fewer FLOPs than nine of the ten baselines. From $512^2$ to $2048^2$, PVT-S shows $20.0\times$ FLOP growth and $16.0\times$ GPU memory growth. For ViT-S/16 and DeiT-S/16, these factors reach $25.3\times$ and $22.5\times$, respectively. DuSAM and the remaining baselines show smaller growth of approximately $15.9$--$16.0\times$ in FLOPs and $9.2$--$10.3\times$ in GPU memory. DuSAM therefore combines low absolute computation with resource growth comparable to the more efficiently scaling baselines.

The measured throughput also supports DuSAM's competitiveness against hierarchical and state-space backbones. At $2048^2$, it is within 4\% of ResNet-50, Swin-T, and ConvNeXt-T, while delivering $1.08\times$, $1.24\times$, $1.32\times$, and $1.95\times$ the throughput of Vim-S, VMamba-T, PVT-S, and PoolFormer-S24, respectively. The mismatch between FLOP and throughput rankings at smaller resolutions motivates the kernel-level analysis in Table~\ref{tab:kernel_profiling}.

\begin{table}[t]
\centering
\caption{Training-time cost of the PM writer on ADE20K. Indented rows decompose the total writer overhead. Shares are relative to the full Standard training step.}
\label{tab:writer_cost}

\setlength{\tabcolsep}{5pt}
\begin{tabular}{@{}lcc@{}}
\toprule

\textbf{Component}
& \textbf{Time (ms)}
& \textbf{Share of step (\%)} \\

\midrule

Training step without writer
& 448.0 & 87.4 \\

\addlinespace[2pt]

\textit{\textbf{Writer overhead (total)}}
& 64.5 & 12.6 \\

\quad Separate stage-4 pass (no PM retrieval)
& 2.0 & 0.4 \\

\quad Region construction and bank commit
& 62.5 & 12.2 \\

\midrule

Full training step (Standard)
& 512.5 & 100.0 \\

\bottomrule
\end{tabular}
\end{table}

\paragraph{Kernel-level Profiling.}
\label{app:Kernel}
As shown in Table~\ref{tab:kernel_profiling}, we use ConvNeXt-T as the profiling reference because it achieves the highest throughput at $512^2$ among the evaluated segmentors. At this resolution, DuSAM launches 1,281 kernels per forward pass versus 259 for ConvNeXt-T. WM and PM operations account for 66\% of launches (51\% and 15\%, respectively) but only 3\% of GPU time, reflecting the use of many short kernels for iterative Sinkhorn balancing, retrieval scoring, top-$r$ selection, and gathering. Together with GPU busy fractions of 56\% versus 96\%, this points to launch and dispatch overhead in the current implementation. At $2048^2$, kernel counts remain nearly constant (1,265 and 263, respectively), while DuSAM's GPU busy fraction rises to 95\% and the memory-operation share of GPU time falls to 2\%. Its throughput deficit relative to ConvNeXt-T narrows from 26.2\% to 3.8\% (Table~\ref{tab:inference_efficiency}). This resolution dependence is consistent with better amortization of launch overhead, motivating fused Sinkhorn and retrieval kernels to improve low-resolution throughput while retaining the memory architecture.

\paragraph{Training-time cost of the writer.}
\label{app:writer_cost}
The writer uses an additional stage-4 pass without PM retrieval to form prototype updates from image evidence (Section~\ref{sec:pm}). To quantify the cost of this design, we compare training steps with the writer disabled, with only its additional stage-4 pass retained, and with the complete writer enabled. Differences between these configurations give the component costs in Table~\ref{tab:writer_cost}. The additional backbone computation is inexpensive: at $512^2$, the separate stage-4 pass processes 256 tokens without gradient tracking and adds only 2.0\,ms (0.4\% of the full training step). The complete writer adds 64.5\,ms (12.6\% of the full step), with the remaining 62.5\,ms attributable to region construction and bank commit. These operations therefore provide the main opportunities for reducing the writer's training overhead. Peak GPU memory is 25.92\,GiB in all three FP32 configurations and 18.98--18.99\,GiB under the FP16 convention of Table~\ref{tab:ade20k_segmentation}. DuSAM thus supports persistent-memory updates with an inexpensive extra backbone pass and nearly unchanged peak training memory. The writer is inactive at inference, so this update mechanism adds no inference-time computation.

\begin{table}[!t]
\centering
\caption{ COCO val2017 results by object size for the models in Table~\ref{tab:coco_detection}. Subscripts $s$, $m$, and $l$ denote small, medium, and large objects. $\dagger$ indicates an additional SimpleFPN is used. $^\star$: stronger augmentation ImageNet-1K pretraining.}
\label{tab:coco_detection_size}
\setlength{\tabcolsep}{1.5pt}
\begin{tabularx}{\linewidth}{@{}>{\raggedright\arraybackslash}Xcc*{6}{c}@{}}
\toprule
\multirow{2}{*}{Method}
& \multirow{2}{*}{Architecture}
& \multirow{2}{*}{\#param.}
& \multicolumn{3}{c}{Bounding box }
& \multicolumn{3}{c}{Instance mask } \\
\cmidrule(lr){4-6}
\cmidrule(lr){7-9}
&
&
& $\mathrm{AP}_{s}$
& $\mathrm{AP}_{m}$
& $\mathrm{AP}_{l}$
& $\mathrm{AP}_{s}$
& $\mathrm{AP}_{m}$
& $\mathrm{AP}_{l}$ \\
\midrule

ResNet-18 \citep{DBLP:conf/cvpr/HeZRS16}
& CNN
& 31.3M
& 19.2 & 36.7 & 43.7
& 14.6 & 33.2 & 44.8 \\

ResNet-34 \citep{DBLP:conf/cvpr/HeZRS16}
& CNN
& 41.2M
& 21.4 & 40.4 & 48.5
& 16.0 & 36.3 & 49.2 \\

ResNet-50 \citep{DBLP:conf/cvpr/HeZRS16}
& CNN
& 44.2M
& 21.9 & 40.9 & 49.5
& 18.3 & 37.4 & 47.2 \\

ResNet-101 \citep{DBLP:conf/cvpr/HeZRS16}
& CNN
& 63.2M
& 22.6 & 44.0 & 52.6
& 18.8 & 39.7 & 49.5 \\

SE-ResNet-50 \citep{DBLP:conf/cvpr/HuSS18}
& CNN
& 46.7M
& 23.4 & 42.7 & 50.0
& 17.1 & 38.6 & 51.8 \\

ConvNeXt-T \citep{DBLP:conf/cvpr/0003MWFDX22}
& CNN
& 48.1M
& 26.5 & 47.4 & 56.8
& 20.4 & 43.0 & 58.8 \\

\midrule

ViT-T $^\dagger$ \citep{DBLP:conf/iclr/DosovitskiyB0WZ21}
& Transformer
& 27.9M
& 19.9 & 37.0 & 48.3
& 15.2 & 34.7 & 51.3 \\

ViT-S $^\dagger$ \citep{DBLP:conf/iclr/DosovitskiyB0WZ21}
& Transformer
& 43.7M
& 23.9 & 41.7 & 54.4
& 17.8 & 38.8 & 55.7 \\

DeiT-S $^\dagger$ \citep{DBLP:conf/icml/TouvronCDMSJ21}
& Transformer
& 43.7M
& 24.2 & 42.4 & 55.6
& 17.8 & 39.0 & 56.6 \\

Swin-T \citep{DBLP:conf/iccv/LiuL00W0LG21}
& Transformer
& 47.8M
& 25.8 & 46.2 & 56.1
& 19.2 & 42.2 & 57.6 \\

PVT-T \citep{DBLP:conf/iccv/WangX0FSLL0021}
& Transformer
& 32.9M
& 21.6 & 39.2 & 49.0
& 19.5 & 37.4 & 48.5 \\

PVT-S \citep{DBLP:conf/iccv/WangX0FSLL0021}
& Transformer
& 44.1M
& 22.9 & 43.0 & 55.4
& 20.4 & 40.3 & 53.6 \\

\midrule

MLP-Mixer-S/16 $^\dagger$ \citep{DBLP:conf/nips/TolstikhinHKBZU21}
& MLP
& 41.6M
& 16.6 & 29.6 & 37.5
& 12.9 & 28.1 & 41.4 \\

Swin-Mixer-T/D24 \citep{DBLP:conf/iccv/LiuL00W0LG21}
& MLP
& 39.9M
& 23.1 & 43.0 & 52.9
& 17.5 & 39.1 & 54.5 \\

\midrule

PoolFormer-S12 \citep{DBLP:conf/cvpr/YuLZSZWFY22}
& MetaFormer
& 31.6M
& 21.4 & 40.2 & 48.9
& 16.0 & 37.1 & 50.9 \\

PoolFormer-S24 \citep{DBLP:conf/cvpr/YuLZSZWFY22}
& MetaFormer
& 41.0M
& 23.6 & 43.5 & 52.4
& 17.7 & 39.8 & 54.0 \\

\midrule

Vim-T $^\dagger$ \citep{DBLP:conf/icml/ZhuL0W0W24}
& SSM
& 27.2M
& 18.0 & 36.1 & 48.5
& 14.6 & 33.3 & 46.3 \\

Vim-S $^\dagger$ \citep{DBLP:conf/icml/ZhuL0W0W24}
& SSM
& 46.3M
& 20.9 & 42.0 & 55.1
& 16.9 & 38.0 & 51.4 \\

MSVMamba-Micro~\citep{DBLP:conf/nips/ShiDX24} 
&  SSM
&  32.0M 
&  26.7  &  47.4  &  58.2 
&  20.4  &  42.9  &  58.3  \\

MambaVision-T~\citep{DBLP:conf/cvpr/HatamizadehK25}  
& SSM+Trans. 
&  51.4M 
&  29.1  &  48.4  &  60.2 
&  22.0  &  44.2  &  61.0 \\

\midrule

\textbf{DuSAM (ours)}
& \textbf{Memory}
& 39.5M
& 26.2 & 45.3 & 55.0
& 19.8 & 40.8 & 55.3 \\

\textbf{DuSAM (ours)$^\star$} 
& \textbf{Memory} 
& 39.5M 
& 26.8  & 46.1  & 55.7 
& 20.7  & 41.9  & 56.2  \\

\bottomrule
\end{tabularx}
\end{table}

\subsection{Additional Experiments and Analysis}
\label{app:memory_analysis}

\subsubsection{COCO Results by Object Size}
\label{app:coco_size}
Table~\ref{tab:coco_detection_size} breaks down the results in Table~\ref{tab:coco_detection} by object size. DuSAM$^\star$ improves both box and mask AP over ResNet-50 and ResNet-101 across all three size groups. Relative to ResNet-50, box AP increases by $4.9$, $5.2$, and $6.2$ for small, medium, and large objects, respectively. The gains also extend to larger-object delineation: large-object mask AP exceeds ResNet-101 by $6.7$. Compared with DeiT-S, DuSAM improves small-object box/mask AP by $2.6/2.9$ and medium-object AP by $3.7/2.9$. Small-object box/mask AP also exceeds Swin-T by $1.0/1.5$ and is the second highest among all compared models, above ConvNeXt-T and MSVMamba-Micro. All six size-specific metrics improve over Swin-Mixer-T/D24 and PoolFormer-S24, while the gains over Vim-S reach $5.9$ box AP and $3.8$ mask AP on small objects. These comparisons show that DuSAM's benefits extend to both small-instance localization and large-object mask prediction, complementing the overall AP results.

\subsubsection{Memory capacity and sensitivity}
\label{app:memory_capacity}

\begin{figure}[t]
    \centering
    \includegraphics[width=\linewidth]{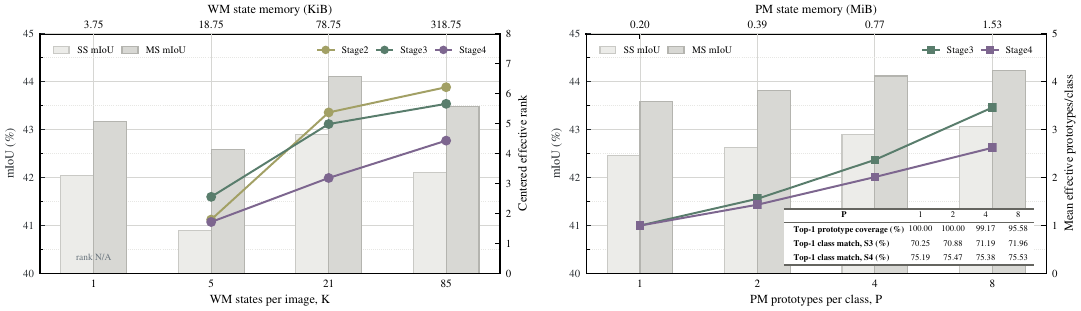}
    \caption{Memory capacity and utilization on ADE20K.}
    \label{fig:memory_capacity}
\end{figure}

\paragraph{WM capacity and state diversity.}
To examine the capacity needed for spatial context, we compare four WM layouts (Fig.~\ref{fig:memory_capacity}, left). Cumulatively adding $1\times1$, $2\times2$, $4\times4$, and $8\times8$ anchor grids gives $K\in\{1,5,21,85\}$. The default $K=21$ achieves the highest SS/MS mIoU with 78.75\,KiB of WM storage per image, approximately one quarter of that required by $K=85$. To assess whether more states capture more varied features, we measure centered effective rank. Its modest increase from $K=21$ to $85$ suggests limited additional feature diversity despite the larger workspace.

\paragraph{PM capacity and utilization.}
To examine the balance between semantic capacity and storage, we vary the number of prototypes per class over $P\in\{1,2,4,8\}$ (Fig.~\ref{fig:memory_capacity}, right). Accuracy improves modestly with capacity. Although $P=8$ performs best, we retain $P=4$ as a compact default with a small accuracy gap and approximately half the PM storage (0.77 versus 1.53\,MiB). At $P=4$, 99.17\% of prototypes are selected as the highest-weight prototype at least once across Stages 3 and 4. The highest-weight prototype matches the query's class in 71.19\% of queries at Stage 3 and 75.38\% at Stage 4. Effective prototype counts indicate broader within-class use at Stage 3, while Stage 4 achieves higher class agreement.

\begin{figure}[!t]
    \centering
    \includegraphics[width=\linewidth]{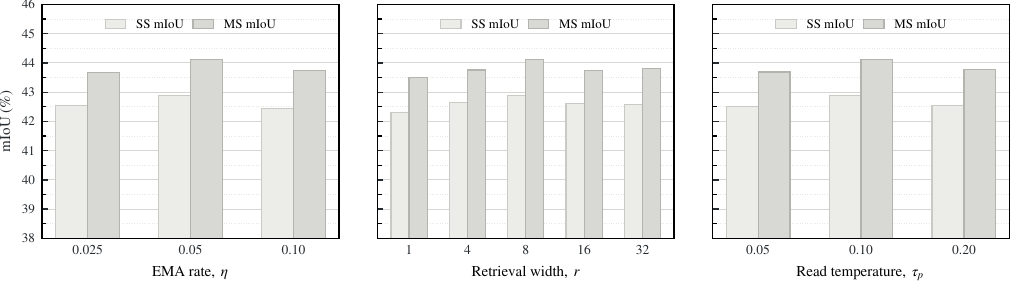}
    \caption{Sensitivity to PM update and retrieval settings. SS/MS mIoU on ADE20K under different EMA rates $\eta$, retrieval widths $r$, and read temperatures $\tau_p$. }
    \label{fig:pm_sensitivity}
\end{figure}

\begin{figure}[t]
    \centering
    \includegraphics[width=\linewidth]{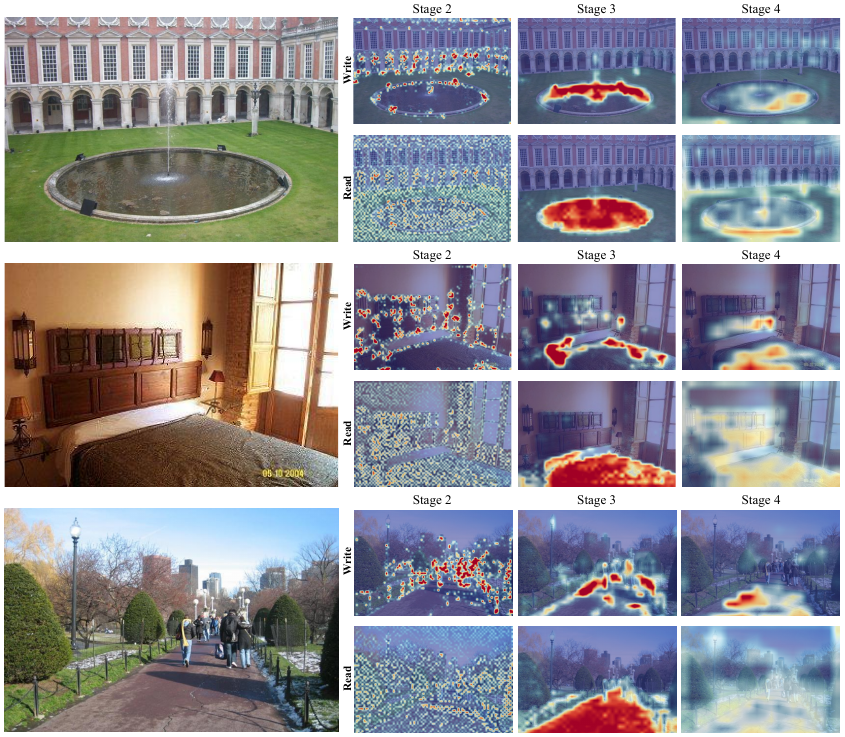}
    \caption{
 Cross-stage WM routing for the same logical state within each example. Columns correspond to Stages~2--4. \emph{Upper:} current write-source density. \emph{Lower:} spatial read weights.
    }
    \label{fig:wm_cross_stage}
\end{figure}

\paragraph{Update and retrieval sensitivity.}
\label{app:pm_sensitivity}
To assess sensitivity to how PM is updated and queried, we vary the EMA update rate $\eta$, retrieval width $r$, and read temperature $\tau_p$ one at a time (Fig.~\ref{fig:pm_sensitivity}). Each configuration is trained for $160$k updates with other settings fixed. The defaults, $\eta=0.05$, $r=8$, and $\tau_p=0.1$, achieve the highest SS/MS mIoU among the tested values. Performance varies modestly with the update rate and temperature. Restricting retrieval to one prototype lowers accuracy, while increasing the retrieval width beyond $r=8$ provides no further gain.

\begin{figure}[t]
    \centering
    \includegraphics[width=\linewidth]{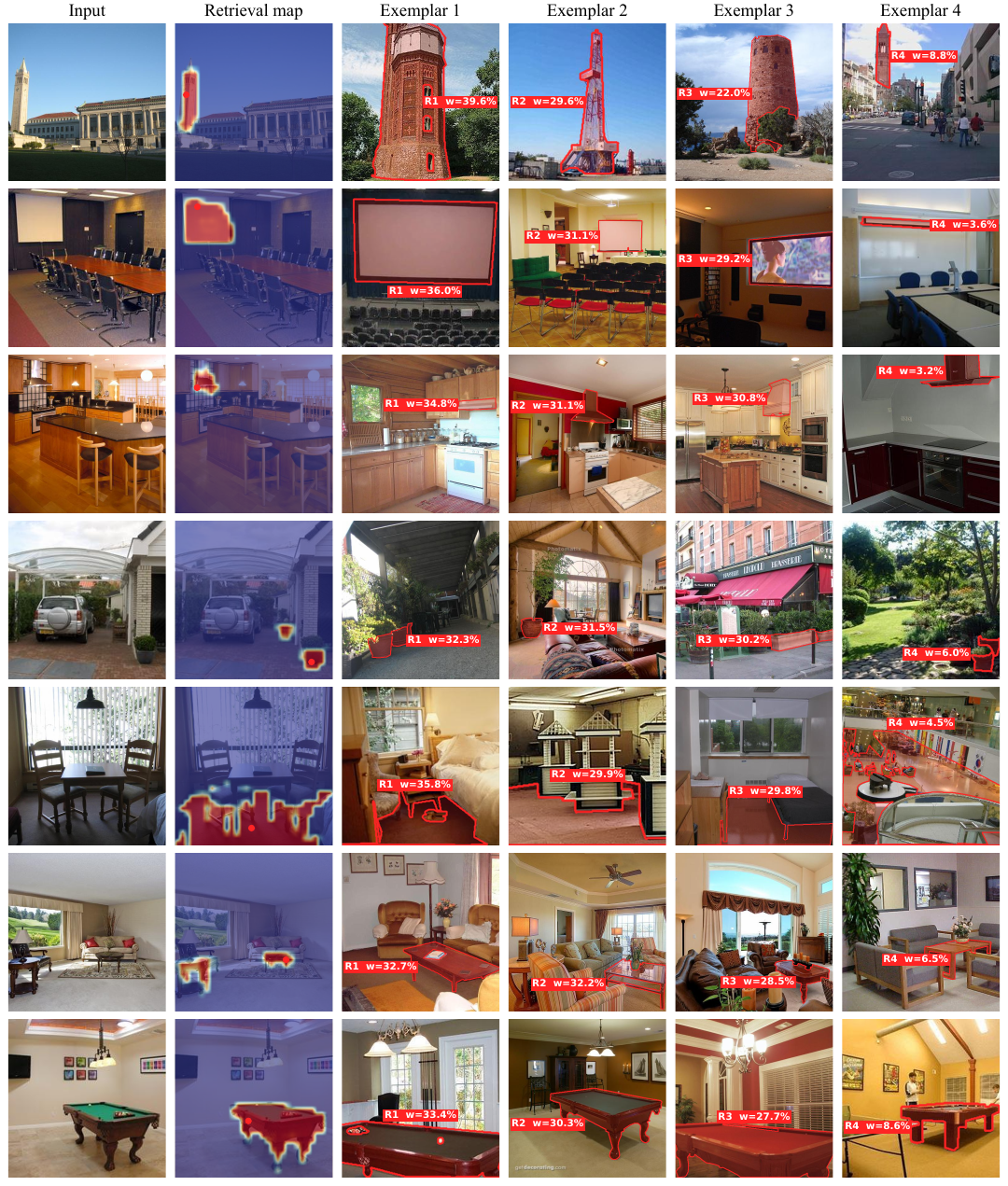}
    \caption{Additional PM retrieval examples. Red outlines mark exemplar regions, $w$ denotes retrieval weights without renormalization.}
    \label{fig:pm_retrieval_more}
\end{figure}

\subsubsection{Cross-Stage Working-Memory Routing}
\label{app:wm_cross_stage}
To examine how working memory is reused through the hierarchy, Fig.~\ref{fig:wm_cross_stage} tracks the same state index $k$ and spatial anchor across Stages~2--4 within each image. Write maps show newly gathered evidence, while read maps show where the updated state is accessed. In these examples, Stage~2 weights are spatially scattered, whereas Stage~3 read maps concentrate on coherent regions such as the fountain, bed, and path. Stage~4 further redistributes access over broader scene areas. These changes illustrate how the workspace adapts both context collection and spatial readback as features evolve, while maintaining correspondence between states across stages. All maps are displayed at image resolution; the coarser feature grids at deeper stages also contribute to their smoother appearance.

\begin{figure}[t]
    \centering
    \includegraphics[width=\linewidth]{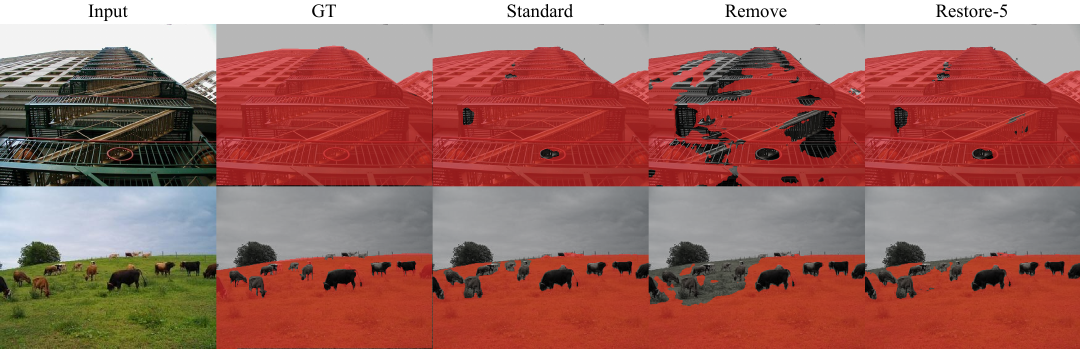}
    \caption{Additional class-targeted PM removal and restoration on ADE20K. Restore-5 uses five annotated training regions with fixed weights.
    }
    \label{fig:pm_restoration_more}
\end{figure}

\begin{figure}[t]
    \centering
    \includegraphics[width=\linewidth]{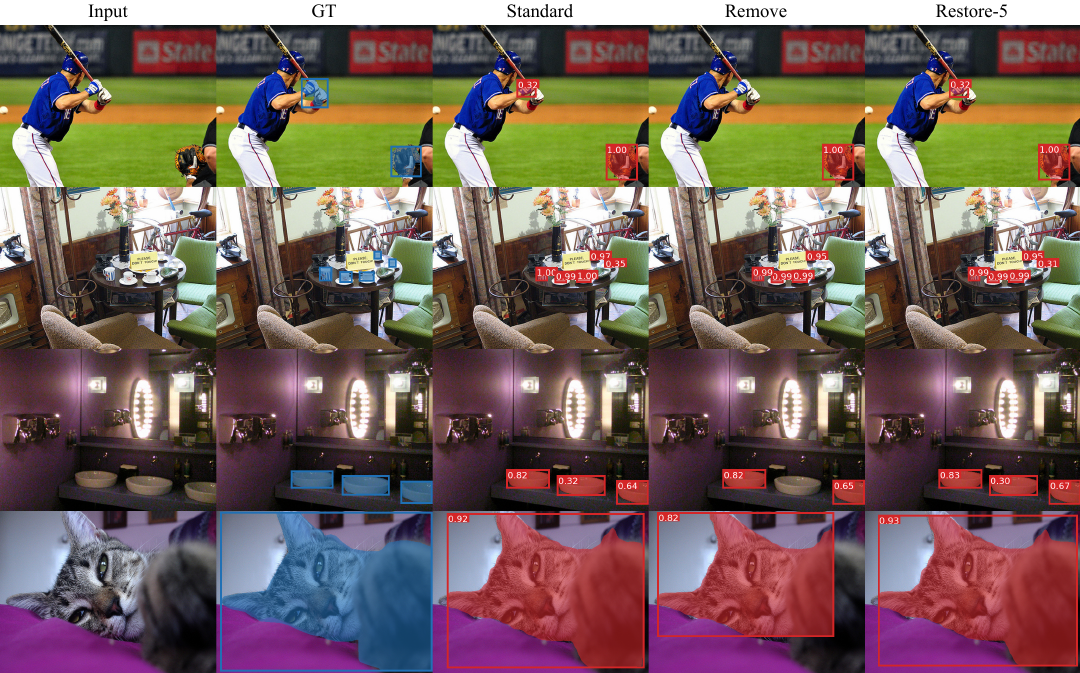}
    \caption{Additional class-targeted PM removal and restoration on COCO. Restore-5 uses five annotated training instances with fixed weights. Blue and red indicate ground truth and predictions.}
    \label{fig:coco_pm_restoration}
\end{figure}

\subsubsection{Additional PM Retrieval Examples}
\label{app:pm_retrieval_examples}
To examine the visual content associated with PM retrieval across categories, Fig.~\ref{fig:pm_retrieval_more} extends Fig.~\ref{fig:pm_retrieval} with additional examples. Heatmaps sum target-class retrieval weights at each image location. Each exemplar is the nearest training region to one of the marked query's four highest-weight prototypes. 

The maps concentrate around localized targets such as range hoods and planters, while floor-related retrieval extends across visible ground surfaces. The associated exemplars also vary within each category, including range hoods of different designs and billiard tables with different colors and viewpoints. Together, these examples illustrate spatially targeted access to category-related experience across varied appearances and scene contexts.

\subsubsection{Additional Class-Targeted PM Removal and Restoration}
\label{app:pm_restoration_examples}

\paragraph{Additional examples.}
Figures~\ref{fig:pm_restoration_more} and~\ref{fig:coco_pm_restoration} extend the class-targeted updates to additional ADE20K and COCO examples. We remove the target class's PM entries and replenish them using five annotated training regions for ADE20K or five training instances for COCO, keeping network parameters fixed. On ADE20K, restoration recovers much of the building and grass coverage lost after removal. On COCO, it restores missing detections in the cup and sink scenes and recovers the cat's predicted extent. These examples show that category-specific memory updates can affect both dense semantic predictions and instance-level localization and masks.

\paragraph{Persistent memory diagnostics.}
To understand how stored semantic evidence affects predictions, we examine cases in which removing a class's PM entries improves its IoU. Table~\ref{tab:negatives} reports the 12 classes with the largest improvements, spanning all four annotated-pixel frequency groups. Each removal is evaluated on the full ADE20K validation set, with network parameters and all other PM entries fixed. Before interpreting these improvements, we check whether they could reflect background fluctuations from modifying PM. For each target class, we therefore use separate removals of the other 149 classes' entries as controls. These leave its own memory intact and provide a reference for how much its IoU changes when other parts of the bank are edited. Against this reference, the gains from removing the target class's own entries are 5--23 times the standard deviation of the control changes.

\begin{table}[t]
\centering
\caption{The classes with the largest IoU gains after removing their own PM entries on ADE20K. Changes are relative to Standard; FP sources identify the ground-truth classes of removed false-positive pixels.}
\label{tab:negatives}

\setlength{\tabcolsep}{3pt}
\renewcommand{\arraystretch}{1.08}

\begin{tabular*}{\linewidth}{@{\extracolsep{\fill}}llrrrll@{}}
\toprule
\multirow{2}{*}{Class}
& \multirow{2}{*}{\shortstack{Frequency\\group}}
& \multirow{2}{*}{\shortstack{$\Delta$IoU\\(pp)}}
& \multirow{2}{*}{\shortstack{$\Delta$TP\\(k px)}}
& \multirow{2}{*}{\shortstack{$\Delta$FP\\(k px)}}
& \multicolumn{2}{c}{Removed FP sources (k px)} \\

\cmidrule(lr){6-7}
& & & & & Largest & Second largest \\

\midrule
grandstand
& High
& $+4.26$
& $-15.0$
& $-105.8$
& stairs~50.1
& wall~19.7 \\

tent
& Lowest
& $+3.23$
& $-0.5$
& $-9.7$
& sky~7.1
& tree~1.2 \\

sea
& Highest
& $+1.63$
& $-107.0$
& $-269.4$
& water~255.6
& animal~3.7 \\

rock
& Highest
& $+1.32$
& $-50.9$
& $-231.1$
& mountain~83.4
& earth~36.3 \\

pier
& Lowest
& $+1.25$
& $-0.6$
& $-3.1$
& wall~1.5
& rock~0.4 \\

bus
& Low
& $+0.96$
& $-2.5$
& $-9.0$
& building~7.3
& truck~0.5 \\

painting
& Highest
& $+0.89$
& $-8.8$
& $-58.0$
& wall~41.5
& poster~4.4 \\

refrigerator
& High
& $+0.88$
& $-8.1$
& $-27.6$
& cabinet~15.1
& wall~6.2 \\

booth
& Low
& $+0.68$
& $-8.1$
& $-31.0$
& building~15.5
& wall~12.2 \\

bed
& Highest
& $+0.66$
& $-138.2$
& $-269.0$
& pillow~43.6
& sofa~43.0 \\

sand
& High
& $+0.54$
& $-13.3$
& $-63.4$
& sea~28.1
& earth~27.0 \\

sink
& High
& $+0.51$
& $-1.5$
& $-10.5$
& bathtub~6.8
& wall~0.9 \\

\bottomrule
\end{tabular*}
\end{table}

To understand why removing these entries improves IoU, we next examine how the target-class predictions change. We compare true positives (TP, correctly recognized target-class pixels) and false positives (FP, pixels of other classes incorrectly predicted as the target) before and after removal. In all 12 cases, both TP and FP decrease. The higher IoU therefore reflects a reduction in false positives that outweighs the loss of correct target predictions. Examining the ground-truth labels of the false positives that disappear reveals several interpretable associations. Shared appearance is a plausible connection for \emph{water} pixels predicted as \emph{sea} and \emph{cabinet} pixels predicted as \emph{refrigerator}. The \emph{tent} case points to a possible role for scene context: the removed false positives include 7.1k \emph{sky} pixels and 1.2k \emph{tree} pixels. For \emph{painting}, \emph{wall} and \emph{poster} pixels connect the affected predictions to a typical supporting surface and a visually related category. The \emph{bed} case combines a component association with \emph{pillows} and shared furniture appearance with \emph{sofas}, while \emph{stairs} and \emph{walls} predicted as \emph{grandstand} suggest an association with constituent architectural structures.

These observations show that the same class memories can support correct recognition while also reinforcing specific confusions with related categories or surrounding structures. A plausible explanation is that some prototypes retain visual or contextual patterns shared with other categories, so retrieving them can strengthen the target label even in regions belonging to another class. Removing the entries suppresses both their useful and misleading contributions; for the reported classes, the reduction in false positives produces a net IoU improvement. This class-specific behavior coexists with the overall benefit of PM shown in Table~\ref{tab:memory_ablation_frequency}. This analysis therefore demonstrates the diagnostic value of DuSAM's explicit memory state: targeted edits allow us to identify the confusions that particular class memories reinforce and the correct predictions they support, without changing network parameters.

\subsubsection{Frozen-feature transfer analysis}
\label{app:frozen_transfer}
We compare fine-tuned and frozen backbones on ADE20K with PM disabled or populated with ADE20K prototypes through gradient-free writes, and additionally evaluate a read-only ImageNet bank in the frozen setting (Table~\ref{tab:frozen_transfer}). WM remains active throughout. With backbone weights frozen, we train the segmentation heads and, for ADE20K PM, the target-label address projection.

With PM disabled, the frozen backbone achieves 36.70/37.92 SS/MS mIoU, showing that its pretrained representations support downstream segmentation without a target-task PM bank. Adding ADE20K prototypes improves SS/MS mIoU by 1.15/1.15 points with backbone fine-tuning, but changes performance by only $+0.12$/$-0.02$ points with a frozen backbone. This contrast suggests that useful target-task memory depends on more than updating its contents and learning new semantic addresses. When the backbone is frozen, its feature transformations and memory readback retain their pretrained parameters, which may limit how effectively newly accumulated evidence is incorporated into features for segmentation. Reusing the read-only ImageNet bank changes SS/MS mIoU by $+0.03$/$+0.39$ points, yielding a near-zero SS change and a small MS change. The limited additional benefit suggests that transferable backbone representations do not automatically make the stored source-task prototypes equally useful for the target task; their category-specific contents may not align with the semantic distinctions needed for ADE20K. Together, these results show that DuSAM learns transferable visual representations, while PM supplies additional task-relevant semantic evidence that improves dense visual prediction after backbone adaptation.

\begin{table}[t]
\centering
\caption{Frozen-feature transfer and PM contribution on ADE20K. We compare frozen and fine-tuned backbones while retaining WM. Parentheses report mIoU changes relative to no PM within each backbone setting.}
\label{tab:frozen_transfer}

\setlength{\tabcolsep}{10pt}
\renewcommand{\arraystretch}{1.10}
\begin{tabular}{@{}lcc@{}}
\toprule
\multicolumn{1}{c}{PM content} & SS mIoU & MS mIoU \\
\midrule
\multicolumn{3}{@{}l}{\textit{\textbf{Fine-tuned backbone}}} \\
\quad No PM
& 41.75 & 42.97 \\
\quad ADE20K prototypes
& 42.90 ($+1.15$) & 44.12 ($+1.15$) \\
\midrule
\multicolumn{3}{@{}l}{\textit{\textbf{Frozen backbone}}} \\
\quad No PM
& 36.70 & 37.92 \\
\quad ADE20K prototypes
& 36.82 ($+0.12$) & 37.90 ($-0.02$) \\
\quad ImageNet prototypes
& 36.73 ($+0.03$) & 38.31 ($+0.39$) \\
\bottomrule
\end{tabular}
\end{table}

\subsection{Limitations and Future Work}
\label{app:limitations}
The current PM is organized by task-specific categories and rebuilt during downstream training, while pretrained backbone weights are transferred. We provide a preliminary analysis of direct reuse of persistent semantic content across label vocabularies. Aligning memory contents and semantic addresses across tasks is a promising direction for retaining this experience during transfer. Our evaluation also focuses on a compact backbone configuration. Extending the study to larger models and more extensive pretraining would help establish how network capacity, memory capacity, and data scale jointly affect performance. These directions would extend the present feasibility study toward more transferable and scalable memory-native visual architectures.

\end{document}